\documentclass{article}

\PassOptionsToPackage{numbers, compress}{natbib}

\usepackage[main, final]{neurips_2026}

\usepackage[utf8]{inputenc}
\usepackage[T1]{fontenc}
\usepackage{hyperref}
\usepackage{url}
\usepackage{booktabs}
\usepackage{amsfonts}
\usepackage{amssymb}
\usepackage{amsmath}
\usepackage{nicefrac}
\usepackage{microtype}
\usepackage[table]{xcolor}
\usepackage{graphicx}
\usepackage{multirow}
\usepackage{wrapfig}
\usepackage{subcaption}

\usepackage{arydshln}
\usepackage{multirow}
\usepackage{booktabs}
\usepackage{float}
\usepackage[table]{xcolor}
\usepackage{array,ragged2e}
\usepackage{bm}
\usepackage{pifont}
\usepackage{amssymb}
\usepackage{amsthm}
\usepackage{algorithm}
\usepackage{algorithmic}
\usepackage{graphicx}
\usepackage[percent]{overpic}
\usepackage{subcaption}
\usepackage{trimclip}
\usepackage{caption}
\usepackage{makecell}
\usepackage{cleveref}
\usepackage{adjustbox}

\newcolumntype{M}[1]{>{\Centering}m{#1}}
\graphicspath{ {./images/} }

\newsavebox{\rightfigbox}

\definecolor{mygreen}{RGB}{0,130,0}

\newif\ifshowcomments
\newif\showpriordraft

\showcommentsfalse 

\ifshowcomments
    \newcommand{\guven}[1]{{\color{orange}#1}}
    \newcommand{\faizan}[1]{{\color{blue}#1}}
    \newcommand{\GD}[1]{{\color{purple}#1}}
    \newcommand{\ES}[1]{{\color{teal}#1}}
    \newcommand{\reza}[1]{{\color{pink}#1}}
    \newcommand{\AG}[1]{{\textbf{\color{cyan}(AG: #1)}}}
    
\else
    \newcommand{\guven}[1]{}
    \newcommand{\faizan}[1]{}
    \newcommand{\GD}[1]{}
    \newcommand{\AG}[1]{}
    \newcommand{\reza}[1]{}
    \newcommand{\GDA}[1]{}
    \newcommand{\ES}[1]{}
\fi
\theoremstyle{remark}

\definecolor{mygray}{gray}{0.4}
\definecolor{mygrayy}{gray}{0.6}
\newcommand{\gy}[1]{{\color{mygray} #1}}
\newcommand{\gyy}[1]{{\color{mygrayy} #1}}

\crefname{figure}{Fig.}{Figs.}
\Crefname{figure}{Fig.}{Figs.}

\crefname{table}{Table}{Tables}
\Crefname{table}{Table}{Tables}

\crefname{section}{Section}{Sections}
\Crefname{section}{Section}{Sections}

\crefname{subsection}{Section}{Sections}
\Crefname{subsection}{Section}{Sections}

\crefname{equation}{Eq.}{Eqs.}
\Crefname{equation}{Eq.}{Eqs.}

\DeclareMathOperator*{\argmin}{arg\,min}

\renewcommand{\arraystretch}{.8}

\definecolor{okblue}{HTML}{0072B2}\definecolor{okvermilion}{HTML}{D55E00}
\definecolor{okorange}{HTML}{E69F00}\definecolor{okpink}{HTML}{CC79A7}

\title{Adaptive Residual Quantization for Memory-Efficient Temporal Action Segmentation}

\author{
  Gerard Donahue$^{1,2\dagger}$ \quad
  Guven Gergerli$^{1,3\dagger}$ \quad
  Ayush Gupta$^{1,4\dagger}$ \\
  \textbf{Reza Ghoddoosian}$^{1}$ \quad
  \textbf{Enna Sachdeva}$^{1}$ \quad
  \textbf{Faizan Siddiqui}$^{1}$ \\[4pt]
  $^1$Honda Research Institute USA \quad $^2$Northeastern University \\
  $^3$Purdue University \quad $^4$Johns Hopkins University \\[2pt]
  {\small\texttt{donahue.g@northeastern.edu}, \texttt{faizan\_siddiqui@honda-ri.com}}
}

\begin{document}

\maketitle
{\renewcommand{\thefootnote}{\fnsymbol{footnote}}\footnotetext[2]{Work done during an internship at Honda Research Institute USA.}}

\begin{abstract}
Temporal action segmentation (TAS) is commonly trained on large banks of pre-extracted frame features, but storing and replaying those features becomes a practical bottleneck for long-form video, especially in incremental TAS where past-task data must remain available to mitigate catastrophic forgetting. Recent TAS replay and condensation methods reduce this burden, but they rely on dense frame-wise annotations and allocate storage through fixed segment-level structures, limiting their flexibility for long procedural videos whose frame complexity varies substantially over time. 
We introduce Adaptive Residual-Quantized Variational AutoEncoder (ARQ-VAE), a compact and label-free memory representation for TAS features that learns a discrete residual quantization of frame-level video features and then adaptively truncates the residual chain on a per-frame basis, retaining only the residual prefix that best reconstructs each frame feature.
This yields a practical compact feature memory that does not depend on frame-wise labels and can therefore support multiple TAS settings with the same stored representation, while a lightweight fine-tuning stage further improves reconstruction quality for the truncated codes that are actually retained. Across standard TAS benchmarks, ARQ-VAE achieves a substantially improved storage--performance trade-off over prior replay and condensation baselines. On Breakfast, our final representation reduces training-set storage from 28~GB for the original I3D feature bank to 10~MB while retaining strong downstream segmentation performance; with ASFormer, it achieves 71.5 Edit and 51.2 F1@50 in the supervised setting. We further show that the same label-free memory remains effective for incremental and unsupervised TAS. We position ARQ-VAE as a practical memory representation for TAS and a promising direction for broader dataset condensation applications.
\end{abstract}

\begin{figure*}[t]
    \centering
    \begin{subfigure}[t]{0.45\linewidth}
        \centering
        \includegraphics[width=\linewidth]{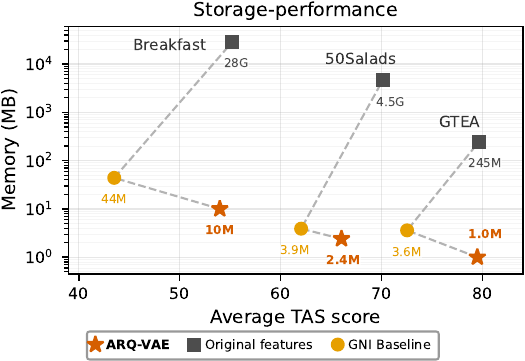}
        \caption{Memory-performance trade-off across datasets.}
        \label{fig:storage-performance}
    \end{subfigure}
    \hfill
    \begin{subfigure}[t]{0.53\linewidth}
        \centering
        \includegraphics[width=\linewidth]{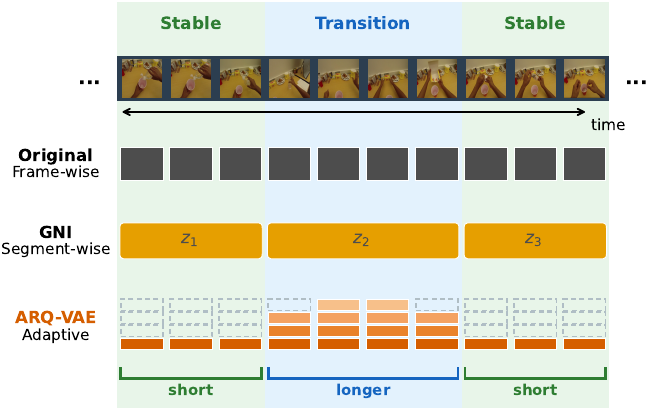}
        \caption{Adaptive memory of ARQ-VAE.}
        \label{fig:transition}
    \end{subfigure}
    \caption{Storage efficiency across TAS settings.
    ARQ-VAE achieves a better storage--performance trade-off while growing memory more slowly than feature replay and GNI.}
    \label{fig:storage_efficiency}
    \vspace{-8pt}
\end{figure*}

\section{Introduction}
Temporal action segmentation (TAS) seeks to assign an action label to every frame in a long untrimmed video, recovering both the temporal extent and ordering of actions over time~\cite{farha2019mstcn,yi2021asformer,ishikawa2021asrf,lu2024fact}. In practice, modern TAS systems are rarely trained directly on raw RGB video.
Instead, they typically rely on pre-extracted frame features from pretrained video backbones~(e.g., I3D~\cite{carreira2017i3d},  DINOv3~\cite{simeoni2025dinov3}, VideoMAE~\cite{tong2022videomae}).
These features provide rich visual-temporal representations and make TAS training more practical, but they also introduce a substantial storage burden: long procedural videos must be stored, loaded, and processed densely at the frame level~\cite{carreira2017i3d,farha2019mstcn,yi2021asformer}.
On benchmarks such as Breakfast and 50Salads, where videos are long and training is frame-wise, feature storage is therefore not merely an implementation detail but a practical bottleneck~\cite{kuehne2014breakfast,stein2013cfiftysalads}.

This bottleneck is particularly sharp in incremental temporal action segmentation (iTAS), where tasks arrive sequentially and models must learn new actions without catastrophically forgetting previously learned ones~\cite{ding2024tca,kirkpatrick2017catastrophic,kemker2018catastrophic,french1999catastrophic}.
Replay-based methods are a natural solution because they recall past task data while learning from new task data~\cite{ding2024tca,iscen2020memory,pellegrini2020latent,liu2020generative,hayes2020remind,wang2021acae}.
However, replay in TAS is expensive: past videos are long, frame-level features are dense, and memory usage grows with each new task. Thus, a compact feature memory is central to whether replay remains practical at all.

Recent TAS-specific methods have begun to address this challenge. TCA~\cite{ding2024tca} reconstructs past features with a conditional generative model trained using action labels and temporal coherence supervision, while GNI~\cite{ding2025gni} stores optimized latent representations that are decoded through a similar label-conditioned generator. Although these methods show that compact replay is effective for TAS, their memory design has two important limitations. First, they require dense frame-wise supervision, restricting them to costly annotation-rich settings. Second, they store coarse action-segment-level latent exemplars, imposing a rigid allocation of memory across time. This is a poor fit for long procedural videos: frame complexity is highly non-uniform, action transitions are often harder than stable interiors, and a useful memory should spend more capacity where the feature sequence is harder to preserve~\cite{kuehne2014breakfast,stein2013cfiftysalads,zhong2024onlinetas,shen2024progress}.

We propose a
compact feature memory for TAS based on residual quantization~(RQ).
RQ represents each frame feature as a sequence of discrete residual codes whose cumulative contribution reconstructs the original continuous feature~\cite{van2017neural,lee2022autoregressive}.
This gives each frame a chain of discrete residual codes that can be stored as a sequence of integer indices.
RQ is attractive for TAS because it provides explicit, compact, supervision-agnostic storage. However, standard RQ is fixed-rate: every frame stores the same number of codes, regardless of each frame's reconstruction difficulty.

Our key idea is to make residual quantization adaptive at the frame level. After training a standard RQ variational autoencoder~(RQ-VAE)~\cite{lee2022autoregressive}, we decode
prefixes of each frame's residual code chain and store an adaptive prefix based on reconstruction behavior.
Unlike fixed-rate or segment-level methods that impose a rigid storage allocation across time, \textbf{Adaptive Residual-Quantized Variational AutoEncoder (ARQ-VAE)} treats reconstruction difficulty as frame-dependent. \Cref{fig:transition} compares our adaptive approach with fixed-rate approaches that operate at the frame and segment levels.
ARQ-VAE retains longer residual prefixes for hard-to-reconstruct frames, such as action transitions or rare visual configurations, and shorter prefixes for redundant frames from stable action interiors.
This produces a compact, label-free feature memory whose capacity adapts to learned per-frame reconstruction complexity. We further introduce a lightweight fine-tuning stage that aligns the decoder and active codebooks with the truncated representations actually stored.

Empirically, as shown in~\cref{fig:storage-performance}, ARQ-VAE improves the storage--performance trade-off across supervised, incremental, and unsupervised TAS settings. On the Breakfast dataset~\cite{kuehne2014breakfast}, ARQ-VAE compresses the original 28~GB I3D feature bank to 10~MB while maintaining strong downstream segmentation performance with standard TAS backbones.
In replay-based iTAS, ARQ-VAE also improves over prior compact baselines, and the same representation generalizes to unsupervised TAS.

In summary, this paper makes three core contributions:
First, we introduce a practical compact memory representation for TAS based on the label-free residual quantization of frame features.
Second, we propose ARQ-VAE, which adaptively allocates residual depth on a per-frame basis
and improves
the storage--performance trade-off relative to fixed-rate quantization.
Third, we demonstrate that this representation is effective across supervised, incremental, and unsupervised TAS.
Together, these results position adaptive residual quantization as a
capable and versatile component for memory-efficient temporal action segmentation.

\section{Related Work}

\paragraph{Temporal action segmentation.}
Temporal action segmentation~(TAS) has been studied under supervised, weakly supervised, unsupervised, and incremental settings. Representative supervised models include MS-TCN, ASRF, ASFormer, and FACT \cite{farha2019mstcn,ishikawa2021asrf,yi2021asformer,lu2024fact}; weaker supervision has been explored through timestamps and activity-level labels \cite{rahaman2022timestamp,ding2022activitylabels}, and unsupervised segmentation through learned representations and clustering \cite{kumar2022unsupervised}. Standard benchmarks include the Breakfast, 50Salads, and GTEA datasets \cite{kuehne2014breakfast,stein2013cfiftysalads,fathi2011lgtea}. More recently, incremental TAS has begun to study sequential task learning under replay constraints. TCA introduces label-conditioned replay for incremental action segmentation \cite{ding2024tca}, and GNI studies the storage--performance trade-off through latent dataset condensation \cite{ding2025gni}. This incremental setting makes memory efficiency especially important, motivating TAS condensation methods that reduce storage while preserving downstream segmentation performance.

\paragraph{Dataset condensation and feature-space replay.}
Dataset distillation and condensation methods reduce memory cost by learning a small synthetic set that substitutes for the original training data \cite{wang2018datasetdistillation,zhao2021datasetcondensation,yang2023efficientdc}. A more closely related line of work compresses the training signal in representation space rather than replaying raw inputs, for example, by storing feature descriptors \cite{iscen2020memory}, replaying intermediate activations \cite{pellegrini2020latent}, synthesizing classifier-side features \cite{liu2020generative}, or quantizing hidden representations for compressed replay \cite{hayes2020remind,wang2021acae}. Recent work has extended these ideas to long-video settings through latent replay with foundation models and compressed code replay \cite{ostapenko2022foundation,mall2025cram}. Our setting differs from both synthetic-set condensation and small-buffer replay: we retain the full training set and reduce memory by replacing each continuous feature vector with a compact discrete code.

\paragraph{Vector quantization for feature compression.}
Vector quantization~\cite{mentzer2024fsq} is a natural mechanism for compressing continuous representations into compact discrete codes. Learned discrete-representation models such as VQ-VAE \cite{van2017neural} and RQ-VAE \cite{lee2022autoregressive} show that high-dimensional features can be encoded as short code sequences while retaining useful structure.
In audio coding, VRVQ~\cite{chae2025variable} varies the residual depth per frame by training an allocator jointly with the codec; ARQ-VAE instead selects each frame's depth post hoc from a standard trained RQ-VAE.
Recent work has expanded this framework with alternative quantization schemes and visual tokenizers, including finite scalar quantization, lookup-free quantization, joint image-video tokenization, and binary spherical quantization \cite{mentzer2024fsq,yu2024magvitv2,wang2024omnitokenizer,zhao2025bsq}, as well as improved methods for learning and updating codebooks \cite{zheng2023online,qin2025learning}. Classical compact-coding methods such as product, additive, and composite quantization are also related because they target efficient coding of continuous vectors for storage and retrieval \cite{jegou2011pq,babenko2014aq,zhang2014cq}. In continual learning, quantization has additionally been used for prototype augmentation and prompt selection \cite{malepathirana2023napavq,jiao2024vqprompt}. In contrast, we use residual quantization specifically as a storage mechanism for fixed pre-extracted video features.

\section{Method}

\subsection{Problem Setup}

\paragraph{Compact Feature Memory for Temporal Action Segmentation.}
Temporal action segmentation (TAS) maps a sequence of pre-extracted frame features
$X=\{x_t\}_{t=1}^{T}$, where $x_t \in \mathbb{R}^{D}$, to a framewise label sequence
$Y=\{y_t\}_{t=1}^{T}$ with $y_t \in \{1,\dots,C\}$, where $C$ is the number of action classes.
We consider the standard feature-based TAS setting used by modern segmentation models, where training and inference operate on dense frame features rather than raw RGB video~\cite{farha2019mstcn,yi2021asformer,lu2024fact}.
Let the TAS training set be
$\mathcal{D}=\{X_i\}_{i=1}^{N}$,
where each training video $i$ is a sequence of $T_i$ frame features
$X_i=\{x_t\}_{t=1}^{T_i}$.

Our goal is to replace the original feature training set with a compact stored representation
$\hat{\mathcal{D}}$
whose storage cost is significantly smaller than that of $\mathcal{D}$ while simultaneously preserving the information content from $\mathcal{D}$ needed for downstream TAS training.
A TAS model is then trained on reconstructed features from $\hat{\mathcal{D}}$ and evaluated on real test features at inference time.
The downstream TAS architecture and loss remain unchanged; only the stored training bank is replaced.

This formulation is motivated most notably by incremental TAS, where prior-task features must remain available for replay~\cite{ding2024tca}.
At the same time, the memory itself should not depend on framewise labels.
Since TAS is also studied under weakly supervised and unsupervised regimes~\cite{rahaman2022timestamp,ding2022activitylabels,xu2024ASOT}, we seek a \emph{label-free} feature memory that can be reused across training settings rather than one tied to costly frame-wise TAS supervision~\cite{ding2025gni}.

\begin{figure*}[t]
    \centering
    \includegraphics[width=\linewidth]{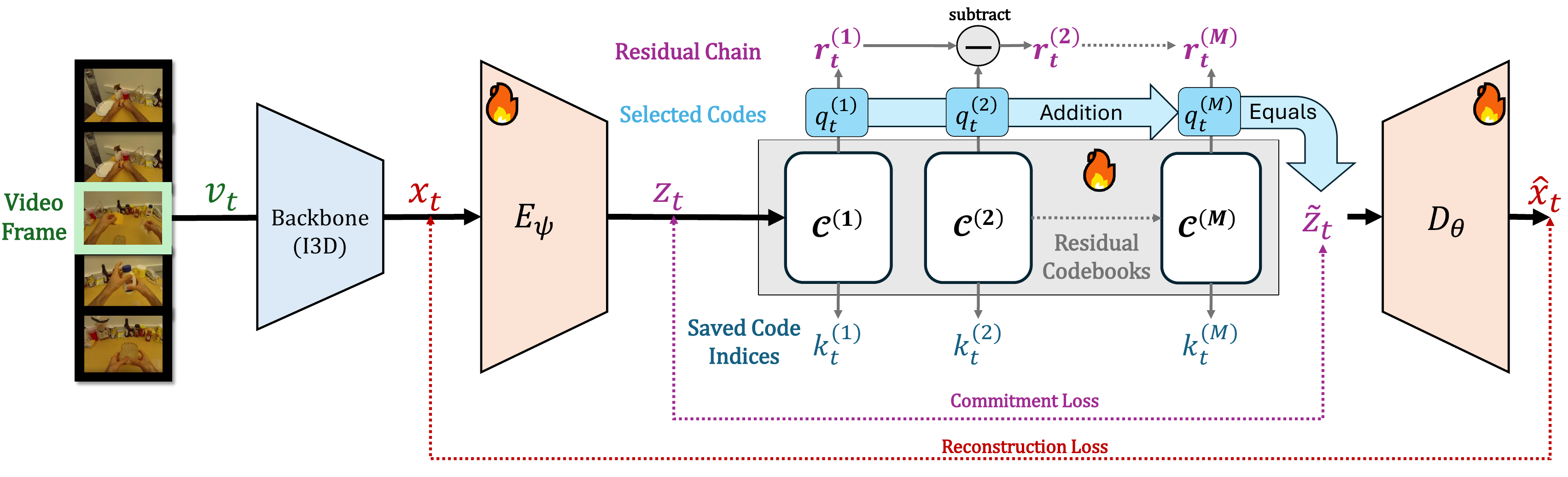}
    \caption{\textbf{Residual-Quantized Variational AutoEncoder (RQ-VAE) for compact TAS feature memory.}
    A pretrained video backbone extracts a frame feature $x_t$ from video frame $v_t$.
    The encoder $E_{\psi}$ maps $x_t$ into latent space, where residual quantization selects a sequence of discrete indices $k_t^{(m)}$ and code vectors $q_t^{(m)}$ from codebooks $\mathcal{C}^{(1)},\dots,\mathcal{C}^{(M)}$.
    These stage-wise codes are summed to form the quantized latent $\tilde z_t=\sum_{m=1}^{M} q_t^{(m)}$, which is decoded by $D_{\theta}$ to reconstruct $\hat x_t$.
    The stored representation is the sequence of discrete stage indices.
    }
    \label{fig:rqvae}
    \vspace{-9pt}
\end{figure*}

\subsection{Residual Quantization for Compact TAS Feature Memory}
\label{sec:RQ-VAE}

We first review the mechanics of residual-quantized variational autoencoders~(RQ-VAE). The core idea of Residual Vector Quantization (RVQ) is to compress high-dimensional continuous vectors by representing them as a cumulative sum of discrete codes. We visualize RQ-VAE in~\cref{fig:rqvae}.

Given a dense frame feature $x_t$ extracted from a video backbone~(e.g., I3D~\cite{carreira2017i3d} or DINOv3~\cite{simeoni2025dinov3}), an encoder $E_{\psi}$ maps it to a continuous latent vector $z_t = E_{\psi}(x_t) \in \mathbb{R}^{d}$. To quantize this vector, RVQ applies a sequence of $M$ refinement steps, which we refer to as \emph{stages}. Each stage $m \in \{1,\dots,M\}$ has a learnable \emph{codebook} $\mathcal{C}^{(m)} \in \mathbb{R}^{K \times d}$, consisting of $K$ embedding vectors $e_k^{(m)} \in \mathbb{R}^d$. Here, $e_k^{(m)}$ denotes the $k$th code vector of stage $m$.

The RVQ process operates iteratively. The first stage ($m=1$) is a coarse discrete approximation of the original latent vector $z_t$. Each subsequent stage then quantizes the error~(or ``residual'')~left over by the previous stage. Specifically, starting with the initial target $r_t^{(1)} = z_t$, each stage $m$ selects the nearest codebook entry to the current residual and computes the new residual for the next stage:
$$k_t^{(m)}=\argmin_{k \in \{1,\dots,K\}} \left\|r_t^{(m)}-e_k^{(m)}\right\|_2^2, \qquad q_t^{(m)}=e_{k_t^{(m)}}^{(m)}, \qquad r_t^{(m+1)} = r_t^{(m)} - q_t^{(m)}.$$
The final quantized latent $\tilde z_t = \tilde z_t^{(M)}$ is the cumulative sum of these selected codes, $\tilde z_t=\sum_{m=1}^{M} q_t^{(m)}$, which a decoder $D_{\theta}$ then maps back to the reconstructed feature $\hat x_t = D_{\theta}(\tilde z_t)$. While utilizing a single stage ($M=1$) reduces this architecture to a standard VQ-VAE~\cite{van2017neural}, the multi-stage residual form provides the higher representational capacity necessary to capture the complex visual-temporal dynamics of long-video features~\cite{lee2022autoregressive}.

As the first step of our pipeline, we train this RQ-VAE at its full, fixed depth $M$ using the standard straight-through estimator~\cite{van2017neural,lee2022autoregressive}:
\begin{equation}
\mathcal{L}_{\mathrm{RQ}}
=
\underbrace{\sum_t \left\|x_t-\hat x_t\right\|_2^2}_{\text{Reconstruction}}
+
\underbrace{\sum_t \sum_{m=1}^{M}
\left(
\left\|\mathrm{sg}[r_t^{(m)}]-q_t^{(m)}\right\|_2^2
+
\beta \left\|r_t^{(m)}-\mathrm{sg}[q_t^{(m)}]\right\|_2^2
\right)}_{\text{Multi-Stage Commitment}},
\label{eq:rq_loss_main}
\end{equation}
where $\mathrm{sg}[\cdot]$ denotes the stop-gradient operation. Intuitively, the multi-stage commitment loss keeps each selected code vector $q_t^{(m)}$ close to the residual $r_t^{(m)}$ it is meant to represent, while also encouraging the encoder residuals to stay close to the selected discrete codes. The stop-gradient operator separates these two updates: the first term updates the selected codes toward the residuals, whereas the second term updates the encoder output to commit to the chosen codes. This separation stabilizes training and helps avoid index collapse, where only a small subset of codebook entries are repeatedly selected~\cite{lee2022autoregressive,gray1984vector}. This corresponds directly to the Commitment Loss depicted in~\cref{fig:rqvae}.

For TAS, RQ-VAE provides a compact, label-free memory by storing only $M$ discrete indices $k_t^{(m)}$ for each frame. However, the standard formulation is strictly \emph{fixed-rate}: every frame utilizes all $M$ stages, regardless of how much capacity individual frames require in practice. We therefore use full-depth RQ-VAE as an initial step and convert it into an adaptive-rate memory described next.

\begin{figure*}[t]
    \centering
    \includegraphics[width=\linewidth]{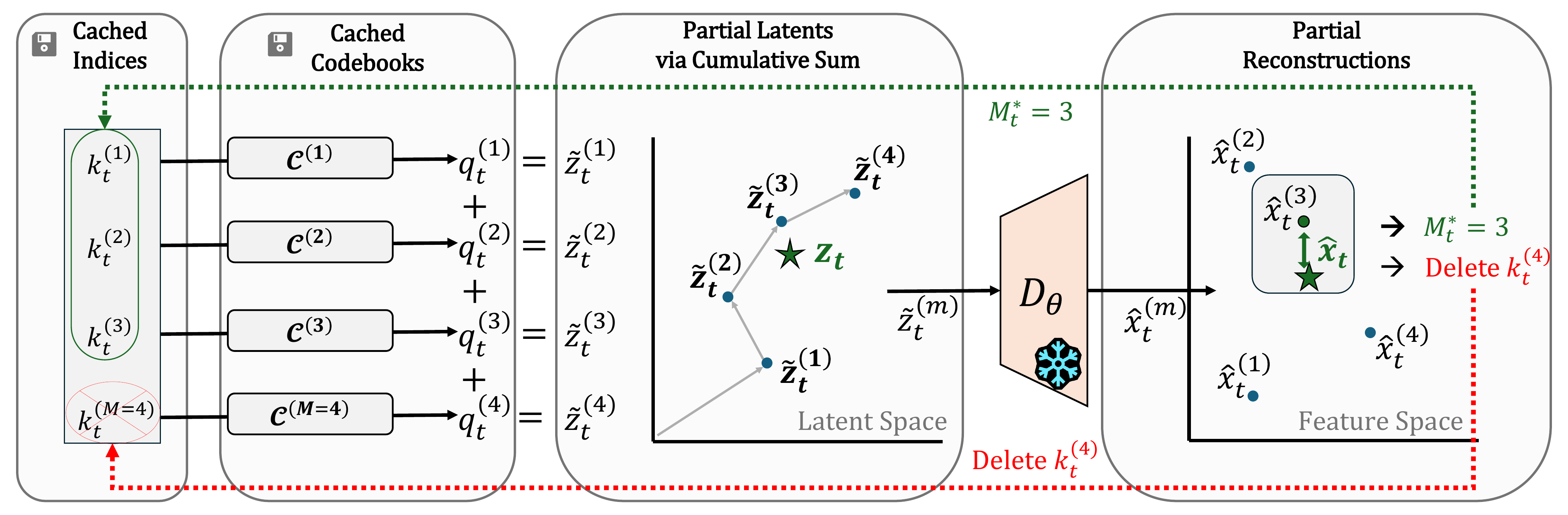}
    \caption{\textbf{Adaptive prefix selection in ARQ-VAE.}
    After training a fixed-depth RQ-VAE, we decode every prefix of the residual chain.
    For frame $t$, the partial latents $\tilde z_t^{(m)}$ yield partial reconstructions $\hat x_t^{(m)}$.
    ARQ-VAE retains the prefix length $m_t^*$ with the lowest reconstruction error~(shown in {\color{teal} green}) and discards the remaining later-stage indices~(shown in {\color{red} red}).
    The resulting memory therefore stores a variable number of stage indices per frame.}
    \label{fig:arqvae}
\end{figure*}

\subsection{Adaptive Residual-Quantized Variational AutoEncoder}
\label{sec:oracle-selection}

To overcome the limitations of fixed-rate residual quantization, we convert the standard RQ-VAE into an adaptive-rate feature memory by selecting, for each frame, the residual-code prefix that minimizes reconstruction error. As illustrated in~\cref{fig:arqvae}, after training the full-depth RQ-VAE, we evaluate every possible \emph{prefix} of the residual chain for a given frame. We define a prefix of length $m$ as the contiguous sequence of residual codes from the first stage up to stage $m$. For a prefix of length $m$, the \textit{partial latent} and \textit{partial reconstruction} are defined as
$$ \tilde z_t^{(m)}=\sum_{i=1}^{m} q_t^{(i)},\quad \hat x_t^{(m)} = D_{\theta}(\tilde z_t^{(m)}). $$

We then select the optimal retained depth for each frame, $m_t^*$, as the prefix length whose partial reconstruction minimizes the L2 error with respect to the original frame feature $x_t$:
\begin{equation}
m_t^*
=
\argmin_{m \in \{1,\dots,M\}}
\left\|x_t-\hat x_t^{(m)}\right\|_2^2.
\label{eq:oracle_prefix_main}
\end{equation}
Consequently, the stored memory for frame $t$ becomes the truncated index sequence $(k_t^{(1)},\dots,k_t^{(m_t^*)})$ rather than the full length-$M$ chain of indices $(k_t^{(1)},\dots,k_t^{(M)})$.

Crucially, this selection rule is entirely data-driven and feature-based: it relies strictly on reconstruction error, requiring no action labels, segment boundaries, or external heuristics. Frames whose features are easily reconstructed by shallow prefixes naturally retain fewer stages. Conversely, frames containing complex visual dynamics that continue to benefit from further refinement retain deeper prefixes (see \cref{sec:downstream-reconstruction} and \cref{fig:recon-stage} for a qualitative analysis demonstrating how stable action interiors are compressed more aggressively than complex action transitions). By allocating storage at the frame level, ARQ-VAE treats feature complexity as highly dynamic, ensuring storage capacity is spent precisely where it empirically improves feature fidelity.

\begin{table}[!t]
    \centering
    \setlength{\tabcolsep}{5pt}
    \renewcommand{\arraystretch}{0.9}
    \caption{\textbf{Temporal Action Segmentation results} on three TAS benchmark datasets.}\vspace{0.3em}
    \resizebox{\linewidth}{!}{
    \begin{tabular}{lcccccccccccc}
    \toprule
    & \multicolumn{4}{c}{GTEA~\cite{fathi2011lgtea}} & \multicolumn{4}{c}{50Salads} & \multicolumn{4}{c}{Breakfast} \\
    \cmidrule(lr){2-5}\cmidrule(lr){6-9}\cmidrule(lr){10-13}
    & Acc & Edit & F1@\{10,25,50\} & Storage
    & Acc & Edit & F1@\{10,25,50\} & Storage
    & Acc & Edit & F1@\{10,25,50\} & Storage \\
    \midrule
    \multicolumn{13}{c}{MS-TCN~\cite{farha2019mstcn}} \\
    \midrule
    \gy{Original}
    & \gy{75.8} & \gy{81.2} & \gy{86.4} / \gy{83.5} / \gy{71.4} & \gy{245 MB}
    & \gy{79.3} & \gy{65.3} & \gy{73.2} / \gy{70.9} / \gy{61.9} & \gy{4.5 GB}
    & \gy{66.5} & \gy{62.8} & \gy{55.7} / \gy{51.0} / \gy{40.1} & \gy{28 GB} \\
    \midrule
    Mean
    & 71.2 & 73.3 & 77.1 / 73.7 / 59.4 & 7.2 MB
    & 69.0 & 42.7 & 50.0 / 46.1 / 37.4 & 7.8 MB
    & 47.6 & 31.8 & 27.8 / 23.3 / 15.6 & 96 MB \\
    Coreset
    & 66.7 & 66.1 & 72.4 / 68.9 / 53.2 & 7.2 MB
    & 61.7 & 43.3 & 49.9 / 46.3 / 35.4 & 7.8 MB
    & 49.7 & 36.8 & 32.3 / 27.5 / 19.3 & 96 MB \\
    TCA~\cite{ding2024tca}
    & 60.9 & 54.1 & 59.2 / 55.3 / 39.3 & --
    & 56.4 & 33.6 & 39.8 / 35.8 / 25.9 & --
    & 34.2 & 20.7 & 17.9 / 13.8 / 8.4 & -- \\
    Encoded
    & 70.4 & 65.5 & 72.2 / 68.8 / 52.1 & 3.6 MB
    & 69.0 & 43.6 & 50.6 / 46.0 / 37.4 & 3.9 MB
    & 37.9 & 49.8 & 40.0 / 32.8 / 19.4 & 44 MB \\
    VQ-VAE
    & 61.5 & 75.1 & 72.0 / 66.2 / 52.5 & 0.1 MB
    & 70.4 & 56.0 & 59.3 / 56.4 / 45.6 & 0.6 MB
    & 49.8 & 48.6 & 40.3 / 34.0 / 22.6 &  4 MB \\
    GNI~\cite{ding2025gni}
    & \textbf{75.2} & 71.9 & 78.3 / 74.6 / 62.7 & 3.6 MB
    & 74.4 & 59.5 & 65.1 / 61.0 / 50.2 & 3.9 MB
    & 55.5 & 45.6 & 46.7 / 41.1 / 28.7 & 44 MB \\
    RQ-VAE
    & 73.5 & 80.6 & 84.6 / 82.2 / 69.1 & 1.1 MB
    & 72.2 & 58.6 & 64.6 / 61.5 / 52.3 & 4.1 MB
    & 58.6 & 57.4 & 51.8 / 45.5 / 33.6 & 22 MB \\
    ARQ-VAE
    & 73.5 & 81.0 & 84.1 / 81.0 / 68.4 & 1.0 MB
    & 73.9 & 60.6 & 64.7 / 61.6 / 52.4 & 2.4 MB
    & 61.8 & 59.4 & 53.8 / 47.9 / 35.8 & 10 MB \\
    ARQ-VAE$^*$
    & 75.1 & \textbf{83.1} & \textbf{86.5} / \textbf{83.0} / \textbf{69.8} & 1.0 MB
    & \textbf{75.5} & \textbf{63.8} & \textbf{69.2} / \textbf{66.0} / \textbf{55.7} & 2.4 MB
    & \textbf{64.6} & \textbf{60.9} & \textbf{55.9} / \textbf{50.3} / \textbf{38.2} & 10 MB \\ \midrule
    \multicolumn{13}{c}{ASFormer~\cite{yi2021asformer}} \\
    \midrule
    \gy{Original}
    & \gy{79.5} & \gy{86.2} & \gy{89.4} / \gy{88.5} / \gy{80.0} & \gy{245 MB}
    & \gy{86.5} & \gy{78.7} & \gy{85.4} / \gy{83.0} / \gy{77.0} & \gy{4.5 GB}
    & \gy{73.4} & \gy{75.6} & \gy{76.7} / \gy{71.3} / \gy{59.0} & \gy{28 GB} \\
    \midrule
    Mean
    & 72.2 & 76.9 & 82.1 / 79.7 / 65.1 & 7.2 MB
    & 71.6 & 49.8 & 56.6 / 52.5 / 43.4 & 7.8 MB
    & 52.2 & 43.2 & 43.5 / 38.3 / 26.7 & 96 MB \\
    Coreset
    & 71.0 & 75.4 & 81.0 / 78.1 / 62.9 & 7.2 MB
    & 69.4 & 46.8 & 56.6 / 52.9 / 39.6 & 7.8 MB
    & 52.0 & 48.3 & 48.3 / 42.4 / 29.7 & 96 MB \\
    TCA~\cite{ding2024tca}
    & 62.2 & 57.8 & 63.0 / 57.4 / 39.9 & --
    & 66.8 & 44.0 & 52.2 / 47.3 / 36.6 & --
    & 36.6 & 28.2 & 26.3 / 22.1 / 14.3 & -- \\
    Encoded
    & 69.2 & 70.2 & 73.3 / 67.3 / 49.8 & 3.6 MB
    & 71.2 & 45.4 & 55.0 / 50.4 / 40.2 & 3.9 MB
    & 37.6 & 53.6 & 50.7 / 41.3 / 24.0 & 44 MB \\
    VQ-VAE
    & 61.3 & 76.4 & 72.8 / 68.4 / 54.9 & 0.1 MB
    & 71.0 & 64.2 & 69.3 / 65.6 / 54.9 & 0.6 MB
    & --- & --- & --- / --- / --- &  4 MB \\
    GNI~\cite{ding2025gni}
    & 77.9 & 82.7 & 86.4 / 84.5 / 70.4 & 3.6 MB
    & \textbf{81.2} & 68.9 & 77.0 / 73.8 / 64.7 & 3.9 MB
    & 59.8 & 48.8 & 54.1 / 47.7 / 34.1 & 44 MB \\
    RQ-VAE
    & 76.9 & 83.5 & 86.2 / 85.2 / 75.3 & 1.1 MB
    & 77.9 & 69.6 & 76.2 / 73.2 / 63.6 & 4.1 MB
    & \textbf{68.9} & 71.4 & \textbf{72.4} / \textbf{65.9} / \textbf{52.9} & 22 MB \\
    ARQ-VAE
    & 77.5 & 82.9 & 87.0 / 85.9 / 73.2 & 1.0 MB
    & 79.2 & 72.2 & 77.3 / 74.5 / 65.6 & 2.4 MB
    & 67.3 & 70.5 & 70.9 / 64.4 / 50.5 & 10 MB \\
    ARQ-VAE$^*$
    & \textbf{78.7} & \textbf{85.3} & \textbf{88.1} / \textbf{86.4} / \textbf{76.6} & 1.0 MB
    & 79.1 & \textbf{72.4} & \textbf{78.8} / \textbf{75.2} / \textbf{65.7} & 2.4 MB
    & 68.3 & \textbf{71.5} & 71.7 / 65.6 / 51.2 & 10 MB \\
    \bottomrule
    \end{tabular}
        } \label{tab:main-tas}\vspace{-0.5em}
\end{table}
\subsubsection{ARQ-VAE Fine-Tuning}
\label{sec:oracle-tuning}

Because prefix selection evaluates truncations using a decoder optimized for full-depth reconstruction, there is a mismatch between the initial training objective and the final stored memory. To resolve this, we introduce a short fine-tuning stage immediately after prefix selection.

After the optimal prefix lengths for each frame $\{m_t^*\}$ are computed, we freeze their corresponding per-frame indices and discard the encoder entirely. We then fine-tune only the decoder $D_{\theta}$ and the codebooks to directly minimize the reconstruction error of the truncated sequences:
\begin{equation}
\mathcal{L}_{\mathrm{ft}}
=
\sum_t \left\|x_t-\hat x_t^{(m_t^*)}\right\|_2^2.
\label{eq:oracle_ft_loss}
\end{equation}
This practical refinement explicitly adapts the network to the precise variable-length representations that will be replayed during downstream TAS training. We use an asterisk (ARQ-VAE$^*$) to refer to the fine-tuned version of ARQ-VAE.

\subsubsection{Downstream Feature Reconstruction and Storage}
\label{sec:downstream-reconstruction}

Once ARQ-VAE$^*$ memory is constructed, the original dense feature bank $\mathcal{D}$ is discarded. The compact representation consists strictly of variable-length index sequences $\{(k_t^{(1)},\dots,k_t^{(m_t^*)})\}_{t=1}^{T_i}$ for all training frames, the stage-wise codebooks, and the fine-tuned decoder $D_{\theta}$. Additionally, we prune the codebooks by discarding any code vectors that are never referenced by the training data, ensuring no storage capacity is wasted on codes never utilized by frames across the dataset.

During downstream TAS training, the network queries this memory rather than reading from a disk-bound feature bank. For a given frame $t$, the reconstructed feature $\hat{x}_t$ is generated on the fly by retrieving the active code vectors corresponding to its stored indices and passing their cumulative sum through $D_{\theta}$. Because the encoder is discarded and the representation relies solely on discrete indices, the overall storage footprint is defined explicitly by the bit-payload of the retained indices ($m_t^* \log_2 K$ bits per frame) and the cost of the pruned floating-point codebook vectors. We present the mathematical breakdown of ARQ-VAE storage accounting in Appendix~\ref{app:storage}.
Following prior work~\cite{ding2025gni}, reported storage excludes the float32 decoder, a fixed 2.369~MB for I3D features regardless of dataset size (Appendix~\ref{app:compression} gives complete footprints and standard compression baselines).

\section{Experiments}\label{sec:experiments}
We evaluate ARQ-VAE across supervised~\cite{farha2019mstcn,yi2021asformer}, incremental~\cite{ding2024tca}, and unsupervised~\cite{xu2024ASOT} TAS, and analyze its behavior through ablations over $K$ and $M$. We report results for ARQ-VAE after adaptive prefix selection~(\ref{sec:oracle-selection}) and ARQ-VAE$^*$ after fine-tuning~(\ref{sec:oracle-tuning}).

\subsection{Experimental Setup}
\paragraph{Datasets.} We utilize four common TAS benchmark datasets: GTEA~\cite{fathi2011lgtea}, 50Salads~\cite{stein2013cfiftysalads}, YouTube Instructional~(YTI)~\cite{alayrac2016YTI}, and Breakfast~\cite{kuehne2014breakfast}. GTEA is a small egocentric dataset with 28 videos and 11 action classes. 50Salads contains 50 long videos with 17 action classes and approximately 20 action instances per video on average. YTI contains instructional videos across 5 activities and is used in the 5-task iTAS setting. Breakfast is the largest benchmark, with 1,712 videos recorded in 18 kitchens and 48 action classes, and is used in the 5- and 10-task iTAS settings. As a result, the storage footprint varies substantially across datasets: GTEA is the smallest, 50Salads and YTI are intermediate, and Breakfast is the largest.

\paragraph{Baselines.}
We compare against TAS replay and condensation baselines from TCA~\cite{ding2024tca} and GNI~\cite{ding2025gni}. \textbf{Original} trains on the full pre-extracted I3D~\cite{carreira2017i3d} feature bank. \textbf{Mean} and \textbf{Coreset} are segment-level baselines: Mean stores the average feature for each ground-truth action segment, while Coreset stores the real frame feature closest to that segment mean; both repeat the stored feature to the original segment length. \textbf{TCA} generates replay features with a label-conditioned temporally coherent VAE, \textbf{Encoded} stores mean TCA encoder latents per segment, and \textbf{GNI} optimizes segment-level latents by network inversion and decodes them with the TCA generator. For incremental TAS, we also report \textbf{Finetune}, which trains on current-task data without replay, and \textbf{Exemplar}, which stores one real frame feature per previous-task segment and repeats it for replay~\cite{ding2024tca}. For unsupervised TAS, we compare reconstructed training features against original I3D features using ASOT~\cite{xu2024ASOT}. See Appendix~\ref{app:baselines} for more details regarding each baseline and its implementation.

\begin{figure*}[!t]
    \centering

    \sbox{\rightfigbox}{\includegraphics[width=0.4\textwidth,height=0.4\textwidth]{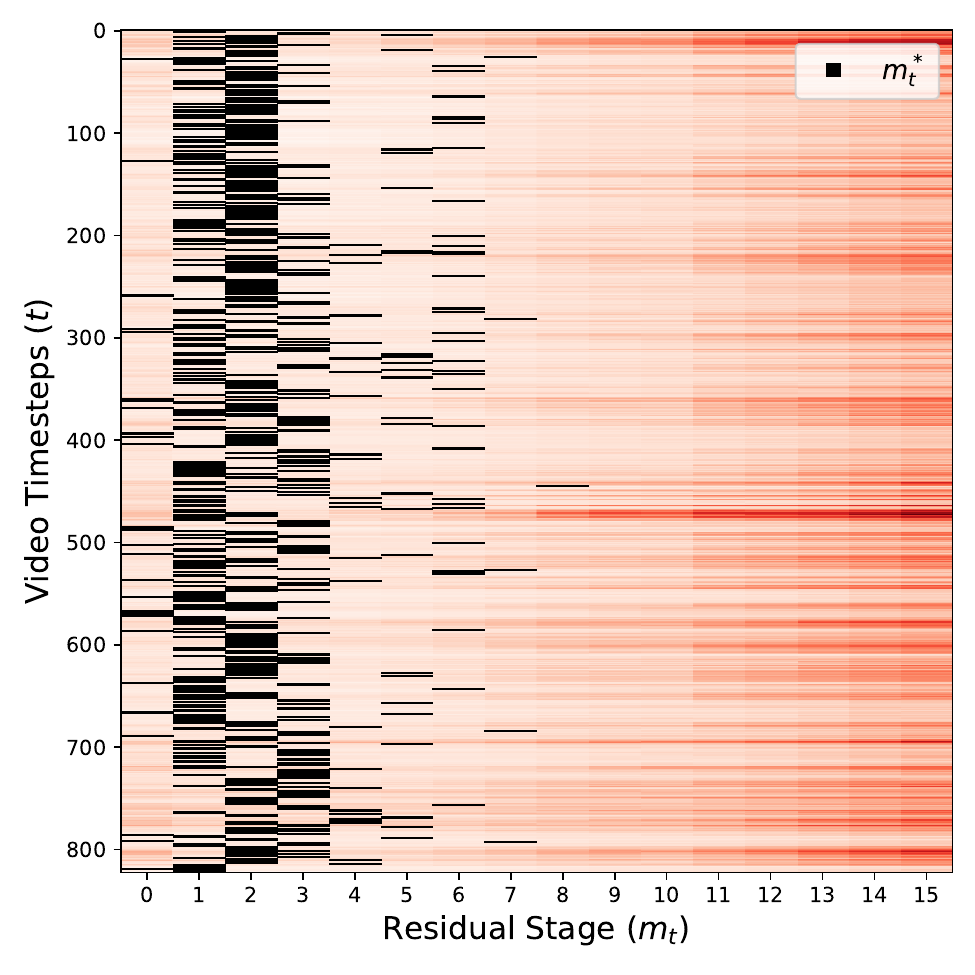}}

    \begin{minipage}[t]{0.58\textwidth}
        \vspace{0pt}
        \captionsetup{type=table}
        \caption{\textbf{Main iTAS results} on Breakfast and YouTube Instructional averaged over 3 task sequences.}
        \label{tab:itas}
        \centering

        \setlength{\tabcolsep}{8pt}
        \renewcommand{\arraystretch}{0.8}

        \adjustbox{max width=\linewidth,
            max totalheight=\dimexpr\ht\rightfigbox+\dp\rightfigbox\relax,
            keepaspectratio
        }{\begin{tabular}{lc|ccccc|ccccc}
        \toprule
        & & \multicolumn{5}{c|}{MS-TCN} & \multicolumn{5}{c}{ASFormer} \\
        \cmidrule(lr){3-7}\cmidrule(lr){8-12}
        \# Tasks & Method & Acc & Edit & \multicolumn{3}{c|}{F1@\{10,25,50\}}
        & Acc & Edit & \multicolumn{3}{c}{F1@\{10,25,50\}} \\
        \midrule
        \multicolumn{12}{c}{Breakfast~\cite{kuehne2014breakfast}} \\
        \midrule

        \multirow{9}{*}{10}
        & \gy{Original}
        & \gy{50.6} & \gy{48.8} & \gy{51.3} & \gy{46.8} & \gy{37.1}
        & \gy{48.1} & \gy{45.2} & \gy{45.9} & \gy{42.4} & \gy{34.2} \\
        & Finetune
        & 7.4 & 7.2 & 7.5 & 7.0 & 5.4
        & 9.9 & 9.8 & 10.3 & 9.4 & 7.5 \\
        & Exemplar
        & 16.1 & 13.3 & 13.8 & 12.5 & 9.5
        & 12.4 & 11.2 & 11.7 & 10.7 & 8.5 \\
        & TCA~\cite{ding2024tca}
        & 29.4 & 25.9 & 26.3 & 23.5 & 17.7
        & 34.2 & 32.4 & 33.1 & 30.1 & 23.4 \\
        & GNI~\cite{ding2025gni}
        & 38.2 & 31.6 & 32.8 & 29.7 & 22.8
        & 46.7 & 41.1 & 41.7 & 38.3 & 30.8 \\
        & RQ-VAE
        & 41.6 & 46.1 & 45.8 & 40.0 & 30.8
        & 40.3 & 44.4 & 45.8 & 40.7 & 31.8 \\
        & ARQ-VAE
        & 49.5 & 51.0 & \textbf{52.1} & \textbf{46.8} & 34.8
        & 53.6 & \textbf{56.1} & \textbf{57.8} & \textbf{51.9} & 42.4 \\
        & ARQ-VAE$^*$
        & \textbf{51.0} & \textbf{51.6} & 51.8 & 46.1 & \textbf{35.0}
        & \textbf{53.9} & 55.7 & 56.9 & 52.4 & \textbf{42.9} \\
        \midrule

        \multirow{9}{*}{5}
        & \gy{Original}
        & \gy{55.8} & \gy{58.0} & \gy{59.9} & \gy{54.5} & \gy{43.5}
        & \gy{60.1} & \gy{59.7} & \gy{63.3} & \gy{56.4} & \gy{45.0} \\
        & Finetune
        & 15.4 & 15.8 & 16.6 & 15.8 & 12.7
        & 15.7 & 16.1 & 16.9 & 15.8 & 13.2 \\
        & Exemplar
        & 32.5 & 28.9 & 30.8 & 28.5 & 22.9
        & 29.5 & 27.5 & 28.7 & 26.7 & 22.0 \\
        & TCA~\cite{ding2024tca}
        & 54.5 & 49.4 & 51.1 & 46.9 & 37.7
        & 57.2 & 56.8 & 58.3 & 54.0 & 43.6 \\
        & GNI~\cite{ding2025gni}
        & 36.5 & 44.5 & 44.6 & 40.4 & 30.9
        & 43.0 & 45.3 & 47.0 & 41.7 & 31.1 \\
        & RQ-VAE
        & 50.5 & 52.8 & 54.6 & 49.1 & 38.3
        & 55.1 & 59.3 & 60.8 & 55.9 & 48.1 \\
        & ARQ-VAE
        & 57.3 & 57.7 & 58.8 & 53.7 & \textbf{42.9}
        & 59.6 & 57.3 & 59.9 & 56.1 & 46.4 \\
        & ARQ-VAE$^*$
        & \textbf{59.0} & \textbf{58.0} & \textbf{59.8} & \textbf{54.2} & 42.5
        & \textbf{64.5} & \textbf{62.9} & \textbf{64.9} & \textbf{60.1} & \textbf{51.8} \\
        \midrule

        \multicolumn{12}{c}{YouTube Instructional~\cite{alayrac2016YTI}} \\
        \midrule

        \multirow{8}{*}{5}
        & \gy{Original}
        & \gy{61.6} & \gy{42.7} & \gy{37.2} & \gy{31.5} & \gy{18.7}
        & \gy{69.0} & \gy{52.9} & \gy{48.7} & \gy{42.7} & \gy{27.6} \\
        & Finetune
        & 13.6 & 2.8 & 3.6 & 2.7 & 0.6
        & 13.9 & 11.5 & 11.1 & 9.8 & 6.3 \\
        & Exemplar
        & 30.8 & 19.7 & 19.8 & 16.0 & 9.3
        & 22.1 & 18.9 & 17.7 & 15.3 & 10.0 \\
        & TCA~\cite{ding2024tca}
        & 30.2 & 25.0 & 21.9 & 18.5 & 11.1
        & 25.2 & 20.9 & 20.1 & 17.5 & 11.4 \\
        & GNI~\cite{ding2025gni}
        & 50.3 & 33.2 & 29.5 & 25.2 & 12.8
        & 53.4 & 36.9 & 31.7 & 25.5 & 12.6 \\
        & RQ-VAE
        & 57.4 & 39.0 & 34.5 & 29.1 & 16.1
        & 63.7 & 46.1 & 42.6 & 37.1 & 21.5 \\
        & ARQ-VAE
        & 63.0 & 44.9 & 40.8 & 34.2 & 19.3
        & \textbf{68.3} & \textbf{52.8} & 47.6 & 40.1 & 24.2 \\
        & ARQ-VAE$^*$
        & \textbf{64.6} & \textbf{48.2} & \textbf{42.1} & \textbf{35.2} & \textbf{20.4}
        & 68.2 & 51.8 & \textbf{47.8} & \textbf{42.1} & \textbf{25.5} \\
        \bottomrule
        \end{tabular}
        }
    \end{minipage}
    \hfill
    \begin{minipage}[t]{0.40\textwidth}
        \caption{Reconstruction MAE heatmap for an example video, showing frame-wise prefix lengths~$m_t^*$ with black boxes.}
        \vspace{0pt}
        \centering
        \usebox{\rightfigbox}
        \captionsetup{type=figure}
        \label{fig:recon-stage}
    \end{minipage}\vspace{-2.4em}
\end{figure*}

\paragraph{Evaluation.} Following prior work~\cite{farha2019mstcn,yi2021asformer}, we report frame-wise Accuracy, segment-wise Edit score, and segment-wise F1@\{10,25,50\} for supervised TAS in~\cref{tab:main-tas} and incremental TAS in~\cref{tab:itas}.
The main supervised TAS results are averaged over the standard dataset splits using $K$-fold cross-validation, while incremental TAS results are averaged over task-wise performance to account for class imbalance and reported across three unique task sequences. For unsupervised TAS in~\cref{tab:asot-table}, we report Mean over Frames~(MoF), F1@50, and mean intersection over union~(mIoU). We also evaluate feature reconstruction quality using Mean Absolute Error~(MAE) and Mean Squared Error~(MSE).

\paragraph{Implementation details.}
Our RQ-VAE follows the lightweight TCA/GNI encoder--decoder design~\cite{ding2024tca,ding2025gni}: two-layer MLP encoder and decoder with latent dimension $d=256$. Residual quantization uses $M=8$ stage-specific codebooks $\{\mathcal{C}^{(m)}\}_{m=1}^{M}$, each in $\mathbb{R}^{K \times d}$ where $K=128$. We train the full-depth RQ-VAE with~\cref{eq:rq_loss_main} for 7.5K iterations using learning rate $1e^{-3}$, then apply prefix selection to store a truncated residual prefix per frame. ARQ-VAE$^*$ fine-tunes for another 7.5K iterations with fixed prefixes, updating only the active codebooks and decoder parameters $\theta$.\footnote{Main experiments used NVIDIA RTX A6000 GPUs with 48~GB memory and CUDA 12.5. The computational-cost study on 50Salads used a V100 GPU (Appendix~\ref{app:compute}). Additional implementation details are in~\cref{app:implementation}.}

For supervised and incremental TAS, we use MS-TCN~\cite{farha2019mstcn} and ASFormer~\cite{yi2021asformer}. MS-TCN is trained for 50 epochs with learning rate $5e^{-4}$, and ASFormer for 120 epochs with learning rate $1e^{-4}$. For unsupervised TAS, we use ASOT~\cite{xu2024ASOT} with reconstructed ARQ-VAE features and follow its original setup. For supervised TAS, ARQ-VAE is trained once on the full training feature bank for each split; for incremental TAS, it is trained separately on each arriving task subset so that only compact previous-task memories are retained for replay.

\begin{table}[!t]
    \centering
    \small
    \setlength{\tabcolsep}{4pt}
    \caption{
        \textbf{Unsupervised action segmentation} with ASOT~\cite{xu2024ASOT}. Gray rows use original I3D features; other rows replace I3D with reconstructed training features.
    }\vspace{0.3em}
    \label{tab:asot-table}
    \resizebox{\textwidth}{!}{\begin{tabular}{l|cccc|cccc|cccc}
        \toprule
        \multirow{2}{*}{Features} &
        \multicolumn{4}{c}{GTEA} &
        \multicolumn{4}{c}{50 Salads} &
        \multicolumn{4}{c}{Breakfast} \\
        \cmidrule(lr){2-5} \cmidrule(lr){6-9} \cmidrule(lr){10-13}
        & MoF & F1 & mIoU & Size & MoF & F1 & mIoU & Size & MoF & F1 & mIoU & Size \\
        \midrule
        \gyy{I3D}
            & \gyy{56.9} & \gyy{52.7} & \gyy{36.1} & \gyy{245 MB} & \gyy{65.2} & \gyy{51.2} & \gyy{32.8} & \gyy{4.5 GB} & \gyy{60.0} & \gyy{58.7} & \gyy{40.0} & \gyy{28 GB} \\
        RQ-VAE
            & 58.1 & 51.9 & 36.8 & 1.3 MB & 60.4 & 43.6 & 27.3 & 4.9 MB & \textbf{65.2} & \textbf{55.0} & \textbf{36.5} & 25 MB \\
        ARQ-VAE
            & \textbf{61.0} & 55.2 & 38.3 & 1.1 MB & 59.1 & 43.4 & 26.1 & 2.7 MB & 64.1 & 53.3 & 34.7 & 11 MB\\
        ARQ-VAE$^*$
            & 59.0 & \textbf{55.9} & \textbf{40.6} & 1.1 MB & \textbf{61.9} & \textbf{45.1} & \textbf{28.0} & 2.7 MB & 64.6 & 54.5 & 36.1 & 11 MB\\
        \bottomrule
    \end{tabular}}\vspace{-1em}
\end{table}

\begin{table}[!t]
    \centering
    \footnotesize
    \setlength{\tabcolsep}{1.7em}
    \caption{\textbf{Summary reconstruction error} and storage averaged across various $(K,M)$ settings.}\vspace{0.3em}
    \label{tab:oracle-summary}
    \begin{tabular}{lccccc}
        \toprule
        \multirow{2}{*}{Dataset}
        & \multicolumn{3}{c}{MSE $\downarrow$}
        & \multicolumn{2}{c}{Storage (MB) $\downarrow$} \\
        \cmidrule(lr){2-4} \cmidrule(lr){5-6}
        & RQ-VAE & ARQ-VAE & ARQ-VAE$^*$ & RQ-VAE & ARQ-VAE \\
        \midrule
        GTEA
        & 0.104
        & 0.026 {\scriptsize \textcolor{mygreen}{(-75.0\%)}}
        & \textbf{0.022} {\scriptsize \textcolor{mygreen}{(-78.8\%)}}
        & 0.989
        & 0.451 {\scriptsize \textcolor{mygreen}{(-54.4\%)}} \\

        50Salads
        & 0.076
        & 0.024 {\scriptsize \textcolor{mygreen}{(-68.4\%)}}
        & \textbf{0.022} {\scriptsize \textcolor{mygreen}{(-70.3\%)}}
        & 4.512
        & 1.471 {\scriptsize \textcolor{mygreen}{(-67.4\%)}} \\

        Breakfast
        & 0.042
        & 0.025 {\scriptsize \textcolor{mygreen}{(-41.5\%)}}
        & \textbf{0.023} {\scriptsize \textcolor{mygreen}{(-45.2\%)}}
        & 21.901
        & 8.392 {\scriptsize \textcolor{mygreen}{(-61.7\%)}} \\
        \bottomrule
    \end{tabular}
\end{table}

\subsection{Experimental Results}
\subsubsection{Supervised Temporal Action Segmentation}
Table~\ref{tab:main-tas} evaluates supervised temporal action segmentation on GTEA~\cite{fathi2011lgtea}, 50Salads~\cite{stein2013cfiftysalads}, and Breakfast~\cite{kuehne2014breakfast} using MS-TCN~\cite{farha2019mstcn} and ASFormer~\cite{yi2021asformer}. We ask whether the proposed discrete memory can preserve segmentation quality while reducing storage relative to the original feature bank and prior compact baselines. Across datasets and backbones, RQ-VAE provides a strong compressed representation, while ARQ-VAE$^*$ achieves a strong overall storage--performance trade-off.

The clearest gains appear on Breakfast, the largest and most storage-intensive benchmark. Compared with GNI, ARQ-VAE$^*$ improves ASFormer from 48.8 to 71.5 Edit and from 34.1 to 51.2 F1@50, while reducing storage from 44~MB to 10~MB. With MS-TCN, it improves Edit from 45.6 to 60.9 and F1@50 from 28.7 to 38.2 under the same storage reduction. These gains suggest that adaptive frame-level allocation is especially beneficial when long videos contain substantial redundancy alongside temporally important frames that require higher reconstruction fidelity. GTEA and 50Salads show the same overall pattern: ARQ-VAE$^*$ closes more of the gap to the original feature bank than prior compact methods, particularly on Edit and F1, indicating better preservation of temporal structure and segment boundaries.

\subsubsection{Incremental Temporal Action Segmentation}
Table~\ref{tab:itas} evaluates replay-based incremental temporal action segmentation under the TCA protocol~\cite{ding2024tca}. Breakfast and YouTube Instructional are partitioned by activity and learned sequentially, with Breakfast evaluated in 10-task and 5-task settings and YouTube Instructional in a 5-task setting. Results are averaged over seen tasks and three unique task sequences.
In this setting, the condensed representation must not only reduce memory cost but also preserve enough information to support replay and mitigate catastrophic forgetting.

Again, the strongest gains appear in the hardest settings, especially on Breakfast. In the 10-task regime, ARQ-VAE and ARQ-VAE$^*$ substantially outperform GNI for both backbones: with MS-TCN, accuracy rises from 38.2 to 49.5 and 51.0, while with ASFormer it rises from 46.7 to 53.6 and 53.9. The segmental metrics follow the same trend. The 5-task Breakfast and 5-task YouTube Instructional results tell a similar story, with our method consistently improving Edit and F1 over prior replay baselines. Together with Table~\ref{tab:main-tas}, these results show that adaptive residual quantization is effective both as a compact training representation and as a substantially stronger replay mechanism for iTAS.

\begin{figure}[t]
    \centering
    \includegraphics[width=\linewidth]{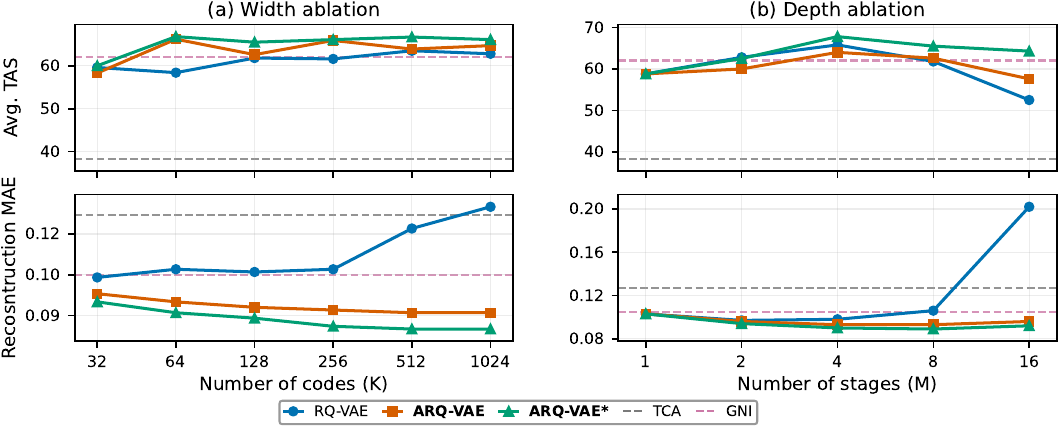}

    \caption{Hyperparameter ablations on the number of codes~($K$) and residual stages~($M$).
    \textbf{(a)} Ablation varying $K$.
    \textbf{(b)} Ablation varying $M$.}

    \label{fig:depth-width-ablation}
    \vspace{-1em}
\end{figure}

\subsubsection{Unsupervised Temporal Action Segmentation}
Table~\ref{tab:asot-table} evaluates ASOT~\cite{xu2024ASOT} when original I3D training features are replaced with reconstructed features. Here, I3D serves as the uncompressed reference, so the goal is to preserve segmentation quality while substantially reducing storage. On GTEA, ARQ-VAE variants match or improve the I3D metrics while reducing storage from 245~MB to 1.1~MB. On 50Salads and Breakfast, the reconstructed features remain reasonably close to I3D while compressing 4.5~GB to 2.7~MB and 28~GB to 11--25~MB, respectively. Overall, these results show that ARQ-VAE remains useful beyond supervised or replay-specific TAS.

\subsubsection{Ablation Studies and Additional Analysis}\label{sec:ablations}

\textbf{Width and depth ablation.} 
We analyze the source of ARQ-VAE's improved storage--performance trade-off through the summary reconstruction and storage results in~\cref{tab:oracle-summary}, the frame-wise prefix visualization in~\cref{fig:recon-stage}, and the width/depth sweeps in~\cref{fig:depth-width-ablation}. \Cref{tab:oracle-summary} shows that adaptive truncation reduces both reconstruction error and storage across \(K \in \{32,64,128,256,512,1024\}\) and \(M \in \{2,4,8,16\}\) (full width/depth sweeps in Appendices~\ref{app:width-ablation}--\ref{app:rate-distortion}). This robustness comes from the prefix selection shown in~\cref{fig:recon-stage}: redundant frames often stop after only a few residual stages, while harder frames retain deeper prefixes. Unlike fixed-depth RQ-VAE, which forces every frame to use the same residual budget even when later stages add little benefit or degrade reconstruction, ARQ-VAE trims unhelpful later stages on a per-frame basis. The sweeps in~\cref{fig:depth-width-ablation} further show that this makes ARQ-VAE less brittle to the choice of codebook width $K$ and residual depth $M$. ARQ-VAE$^*$ performs best overall, indicating that adaptive truncation and fine-tuning together produce a more reliable compact feature memory. 

\begin{figure}[t]
    \centering
    \begin{minipage}[c]{0.46\linewidth}
        \centering
        \includegraphics[width=\linewidth]{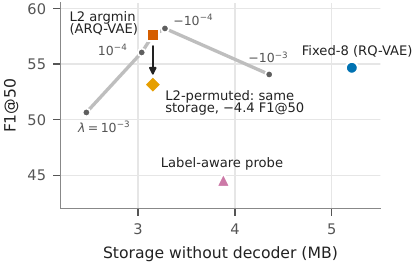}
    \end{minipage}\hfill
    \begin{minipage}[c]{0.52\linewidth}
        \centering
        \setlength{\tabcolsep}{3pt}
        \renewcommand{\arraystretch}{1.45}
        \resizebox{\linewidth}{!}{\begin{tabular}{lcccccc}
            \toprule
            Criterion & MB & Acc & Edit & F1@10 & F1@25 & F1@50 \\
            \midrule
            \textcolor{okblue}{$\bullet$} Fixed-8 & 5.21 & 69.84 & 59.87 & 67.88 & 63.78 & 54.67 \\
            \textcolor{okvermilion}{\tiny$\blacksquare$} L2 argmin & \textbf{3.15} & \textbf{74.92} & \textbf{62.01} & \textbf{69.20} & \textbf{65.18} & \textbf{57.59} \\
            \textcolor{okorange}{\scriptsize$\blacklozenge$} L2-permuted & \textbf{3.15} & 70.50 & 58.98 & 64.41 & 60.81 & 53.15 \\
            \textcolor{okpink}{\scriptsize$\blacktriangle$} Label-aware probe & 3.88 & 68.45 & 51.11 & 57.52 & 54.10 & 44.57 \\
            \bottomrule
        \end{tabular}}
    \end{minipage}
    \caption{\textbf{Prefix-selection criteria} on a single 50Salads split with MS-TCN. \textbf{Left:}~F1@50 vs.\ storage without decoder; the gray curve sweeps $\lambda$ ($\lambda{=}0$ is L2 argmin). \textbf{Right:}~TAS metrics per criterion; MB is storage without decoder. Full results in~\cref{tab:selection-criteria,tab:rd-sweep}.}
    \label{fig:selection}
    \vspace{-0.8em}
\end{figure}

\phantomsection\label{sec:selection}
\textbf{Reconstruction error as a selection criterion.} ARQ-VAE selects prefix lengths by minimizing feature-space reconstruction error~(\cref{eq:oracle_prefix_main}), whereas segmentation also depends on action boundaries and temporal structure. To test whether this choice matters, we reuse the same RQ-VAE so that only the retained prefix lengths differ (\cref{fig:selection}; Appendix~\ref{app:selection}). Unlike the cross-validated results in~\cref{tab:main-tas}, these diagnostics use a single 50Salads train/test split, so their absolute values are not directly comparable. Besides Fixed-8 (RQ-VAE) and L2 argmin (ARQ-VAE), we evaluate \textbf{L2-permuted}, which shuffles the L2-selected depths across frames, preserving the depth histogram and storage while breaking the link between frames and depths, and a supervised \textbf{label-aware probe}, which keeps the prefix with the lowest correct-class cross-entropy under a linear frame-wise classifier trained on TAS labels. At identical storage, shuffling lowers F1@50 from 57.59 to 53.15, below even Fixed-8: frame-specific depth allocation matters beyond the total number of retained codes. The probe uses more storage yet reaches only 44.57 F1@50, so this label-aware probe does not improve selection in this experiment.

\textbf{Explicit rate--distortion control.} L2 argmin does not penalize storage. Selecting $m_t(\lambda)=\argmin_{m}[D_t(m)+\lambda m]$, where $D_t(m)$ is the MSE of the length-$m$ prefix, makes this trade-off explicit without retraining and recovers L2 argmin at $\lambda{=}0$. As the gray curve in~\cref{fig:selection} shows, $\lambda>0$ reduces storage at the expected cost in segmentation performance, a mild bias toward longer prefixes ($\lambda{=}{-}10^{-4}$) slightly improves F1@50 to 58.21, and a strong bias ($\lambda{=}{-}10^{-3}$) lowers it to 54.07, so more stages are not monotonically better. Because RQ-VAE is trained for full-depth reconstruction and its latent residuals pass through a nonlinear decoder, later stages can move an already accurate reconstruction away from its target, even though training shows no sign of codebook collapse. These results support L2 as an effective, if not universal, label-free selection criterion in this setting.

Additional analyses in the appendix include prefix-length histograms~(\ref{app:histo}), empirical frame difficulty~(\ref{app:prefix_difficulty}), qualitative alignment of residual codes with action segments~(\ref{app:code-semantics}), ARQ-VAE performance on contemporary DINOv3 features~(\ref{app:dino}), and computational cost~(\ref{app:compute}).
Notably, prefix selection adds 16.47~s to RQ-VAE preparation time (102.91~s) on the \textbf{50Salads} dataset. For 50 MS-TCN epochs, training with on-the-fly decoding took 212.30~s versus 259.43~s for FP32; while the initial ARQ-VAE run took 331.68~s including one-time preparation and training.

\section{Conclusion}
We presented ARQ-VAE, a compact, label-free feature memory for temporal action segmentation that keeps only the residual prefix that best reconstructs each frame. This allocation improves the storage--performance trade-off over prior replay and condensation methods across supervised, incremental, and unsupervised temporal action segmentation, compressing the 28~GB Breakfast I3D feature bank to 10~MB, and carries over to DINOv3 features and the FACT~\cite{lu2024fact} architecture (Appendices~\ref{app:dino} and~\ref{app:fact}) at a small one-time cost.

\paragraph{Limitations and future work.}
ARQ-VAE compresses pre-extracted features, therefore its effectiveness is bounded by the features of the underlying video backbone.
Additionally, prefix selection in ARQ-VAE relies on a label-free reconstruction proxy. A worthy future direction is to investigate the potential of a prefix selection specific to temporal action segmentation, which could better specialize the adapted codebooks for the TAS setting.
Finally, although the mechanism is task-agnostic, our evidence is specific to TAS; extending adaptive residual memories to other frozen feature caches, such as the latents used to train video diffusion models, is a promising direction for future work.

\bibliographystyle{plainnat}

\bibliography{main}

@inproceedings{lee2022autoregressive,
  title={Autoregressive image generation using residual quantization},
  author={Lee, Doyup and Kim, Chiheon and Kim, Saehoon and Cho, Minsu and Han, Wook-Shin},
  booktitle={Proceedings of the IEEE/CVF conference on computer vision and pattern recognition},
  pages={11523--11532},
  year={2022}
}

@article{van2017neural,
  title={Neural discrete representation learning},
  author={Van Den Oord, Aaron and Vinyals, Oriol and others},
  journal={Advances in neural information processing systems},
  volume={30},
  year={2017}
}

@inproceedings{farha2019mstcn,
  title={Ms-tcn: Multi-stage temporal convolutional network for action segmentation},
  author={Farha, Yazan Abu and Gall, Jurgen},
  booktitle={Proceedings of the IEEE/CVF conference on computer vision and pattern recognition},
  pages={3575--3584},
  year={2019}
}

@inproceedings{fathi2011lgtea,
  title={Learning to recognize objects in egocentric activities},
  author={Fathi, Alireza and Ren, Xiaofeng and Rehg, James M},
  booktitle={CVPR 2011},
  pages={3281--3288},
  year={2011},
  organization={IEEE}
}

@inproceedings{stein2013cfiftysalads,
  title={Combining embedded accelerometers with computer vision for recognizing food preparation activities},
  author={Stein, Sebastian and McKenna, Stephen J},
  booktitle={Proceedings of the 2013 ACM international joint conference on Pervasive and ubiquitous computing},
  pages={729--738},
  year={2013}
}

@inproceedings{kuehne2014breakfast,
  title={The language of actions: Recovering the syntax and semantics of goal-directed human activities},
  author={Kuehne, Hilde and Arslan, Ali and Serre, Thomas},
  booktitle={Proceedings of the IEEE conference on computer vision and pattern recognition},
  pages={780--787},
  year={2014}
}

@inproceedings{carreira2017i3d,
  title={Quo vadis, action recognition? a new model and the kinetics dataset},
  author={Carreira, Joao and Zisserman, Andrew},
  booktitle={proceedings of the IEEE Conference on Computer Vision and Pattern Recognition},
  pages={6299--6308},
  year={2017}
}

@article{tong2022videomae,
  title={Videomae: Masked autoencoders are data-efficient learners for self-supervised video pre-training},
  author={Tong, Zhan and Song, Yibing and Wang, Jue and Wang, Limin},
  journal={Advances in neural information processing systems},
  volume={35},
  pages={10078--10093},
  year={2022}
}

@article{yi2021asformer,
  title={Asformer: Transformer for action segmentation},
  author={Yi, Fangqiu and Wen, Hongyu and Jiang, Tingting},
  journal={arXiv preprint arXiv:2110.08568},
  year={2021}
}

@article{zhong2024onlinetas,
  title={Onlinetas: An online baseline for temporal action segmentation},
  author={Zhong, Qing and Ding, Guodong and Yao, Angela},
  journal={Advances in Neural Information Processing Systems},
  volume={37},
  pages={58984--59005},
  year={2024}
}

@inproceedings{shen2024progress,
  title={Progress-aware online action segmentation for egocentric procedural task videos},
  author={Shen, Yuhan and Elhamifar, Ehsan},
  booktitle={Proceedings of the IEEE/CVF Conference on Computer Vision and Pattern Recognition},
  pages={18186--18197},
  year={2024}
}

@article{gray1984vector,
  title={Vector quantization},
  author={Gray, Robert},
  journal={IEEE Assp Magazine},
  volume={1},
  number={2},
  pages={4--29},
  year={1984},
  publisher={IEEE}
}

@inproceedings{alayrac2016YTI,
  title={Unsupervised learning from narrated instruction videos},
  author={Alayrac, Jean-Baptiste and Bojanowski, Piotr and Agrawal, Nishant and Sivic, Josef and Laptev, Ivan and Lacoste-Julien, Simon},
  booktitle={Proceedings of the IEEE conference on computer vision and pattern recognition},
  pages={4575--4583},
  year={2016}
}

@inproceedings{xu2024ASOT,
  title={Temporally consistent unbalanced optimal transport for unsupervised action segmentation},
  author={Xu, Ming and Gould, Stephen},
  booktitle={Proceedings of the IEEE/CVF Conference on Computer Vision and Pattern Recognition},
  pages={14618--14627},
  year={2024}
}

@article{kirkpatrick2017catastrophic,
  title={Overcoming catastrophic forgetting in neural networks},
  author={Kirkpatrick, James and Pascanu, Razvan and Rabinowitz, Neil and Veness, Joel and Desjardins, Guillaume and Rusu, Andrei A and Milan, Kieran and Quan, John and Ramalho, Tiago and Grabska-Barwinska, Agnieszka and others},
  journal={Proceedings of the national academy of sciences},
  volume={114},
  number={13},
  pages={3521--3526},
  year={2017},
  publisher={National Academy of Sciences}
}

@inproceedings{kemker2018catastrophic,
  title={Measuring catastrophic forgetting in neural networks},
  author={Kemker, Ronald and McClure, Marc and Abitino, Angelina and Hayes, Tyler and Kanan, Christopher},
  booktitle={Proceedings of the AAAI conference on artificial intelligence},
  volume={32},
  number={1},
  year={2018}
}

@article{french1999catastrophic,
  title={Catastrophic forgetting in connectionist networks},
  author={French, Robert M},
  journal={Trends in cognitive sciences},
  volume={3},
  number={4},
  pages={128--135},
  year={1999},
  publisher={Elsevier}
}

@article{wang2018datasetdistillation,
  title={Dataset Distillation},
  author={Wang, Tongzhou and Zhu, Jun-Yan and Torralba, Antonio and Efros, Alexei A.},
  journal={arXiv preprint arXiv:1811.10959},
  year={2018}
}

@inproceedings{zhao2021datasetcondensation,
  title={Dataset Condensation with Gradient Matching},
  author={Zhao, Bo and Mopuri, Konda Reddy and Bilen, Hakan},
  booktitle={International Conference on Learning Representations},
  year={2021},
  url={https://openreview.net/forum?id=mSAKhLYLSsl}
}

@inproceedings{yang2023efficientdc,
  title={An Efficient Dataset Condensation Plugin and Its Application to Continual Learning},
  author={Yang, Enneng and Shen, Li and Wang, Zhenyi and Liu, Tongliang and Guo, Guibing},
  booktitle={Advances in Neural Information Processing Systems},
  volume={36},
  pages={67625--67642},
  year={2023}
}

@inproceedings{iscen2020memory,
  title={Memory-Efficient Incremental Learning Through Feature Adaptation},
  author={Iscen, Ahmet and Zhang, Jeffrey and Lazebnik, Svetlana and Schmid, Cordelia},
  booktitle={Computer Vision -- ECCV 2020},
  series={Lecture Notes in Computer Science},
  volume={12361},
  pages={699--715},
  year={2020},
  doi={10.1007/978-3-030-58517-4_41}
}

@inproceedings{pellegrini2020latent,
  title={Latent Replay for Real-Time Continual Learning},
  author={Pellegrini, Lorenzo and Graffieti, Gabriele and Lomonaco, Vincenzo and Maltoni, Davide},
  booktitle={2020 IEEE/RSJ International Conference on Intelligent Robots and Systems (IROS)},
  pages={10203--10209},
  year={2020},
  doi={10.1109/IROS45743.2020.9341460}
}

@inproceedings{liu2020generative,
  title={Generative Feature Replay For Class-Incremental Learning},
  author={Liu, Xialei and Wu, Chenshen and Menta, Mikel and Herranz, Luis and Raducanu, Bogdan and Bagdanov, Andrew D. and Jui, Shangling and van de Weijer, Joost},
  booktitle={Proceedings of the IEEE/CVF Conference on Computer Vision and Pattern Recognition Workshops},
  pages={915--924},
  year={2020},
  doi={10.1109/CVPRW50498.2020.00121}
}

@inproceedings{hayes2020remind,
  title={REMIND Your Neural Network to Prevent Catastrophic Forgetting},
  author={Hayes, Tyler L. and Kafle, Kushal and Shrestha, Robik and Acharya, Manoj and Kanan, Christopher},
  booktitle={Computer Vision -- ECCV 2020},
  series={Lecture Notes in Computer Science},
  volume={12353},
  pages={466--483},
  year={2020},
  doi={10.1007/978-3-030-58598-3_28}
}

@article{wang2021acae,
  title={ACAE-REMIND for Online Continual Learning with Compressed Feature Replay},
  author={Wang, Kai and van de Weijer, Joost and Herranz, Luis},
  journal={Pattern Recognition Letters},
  volume={150},
  pages={122--129},
  year={2021},
  doi={10.1016/j.patrec.2021.06.025}
}

@inproceedings{ostapenko2022foundation,
  title={Continual Learning with Foundation Models: An Empirical Study of Latent Replay},
  author={Ostapenko, Oleksiy and Lesort, Timoth{\'e}e and Rodr{\'\i}guez, Pau and Arefin, Md Rifat and Douillard, Arthur and Rish, Irina and Charlin, Laurent},
  booktitle={Proceedings of The 1st Conference on Lifelong Learning Agents},
  series={Proceedings of Machine Learning Research},
  volume={199},
  pages={60--91},
  year={2022}
}

@inproceedings{mall2025cram,
  title={CRAM: Large-scale Video Continual Learning with Bootstrapped Compression},
  author={Mall, Shivani and Henriques, Jo{\~a}o F.},
  booktitle={Proceedings of the IEEE/CVF International Conference on Computer Vision},
  pages={15045--15055},
  year={2025}
}

@inproceedings{mentzer2024fsq,
  title     = {Finite Scalar Quantization: {VQ}-{VAE} Made Simple},
  author    = {Mentzer, Fabian and Minnen, David and Agustsson, Eirikur and Tschannen, Michael},
  booktitle = {The Twelfth International Conference on Learning Representations},
  year      = {2024},
  url       = {https://openreview.net/forum?id=8ishA3LxN8}
}

@inproceedings{zheng2023online,
  title     = {Online Clustered Codebook},
  author    = {Zheng, Chuanxia and Vedaldi, Andrea},
  booktitle = {Proceedings of the IEEE/CVF International Conference on Computer Vision},
  year      = {2023},
  doi       = {10.1109/ICCV51070.2023.02084},
  url       = {https://openaccess.thecvf.com/content/ICCV2023/html/Zheng_Online_Clustered_Codebook_ICCV_2023_paper.html}
}

@inproceedings{qin2025learning,
  title     = {Learning to Quantize for Training Vector-Quantized Networks},
  author    = {Qin, Peijia and Zhang, Jianguo},
  booktitle = {Proceedings of the 42nd International Conference on Machine Learning},
  series    = {Proceedings of Machine Learning Research},
  volume    = {267},
  year      = {2025},
  url       = {https://proceedings.mlr.press/v267/qin25j.html}
}

@article{jegou2011pq,
  title   = {Product Quantization for Nearest Neighbor Search},
  author  = {J{\'e}gou, Herv{\'e} and Douze, Matthijs and Schmid, Cordelia},
  journal = {IEEE Transactions on Pattern Analysis and Machine Intelligence},
  volume  = {33},
  number  = {1},
  pages   = {117--128},
  year    = {2011},
  doi     = {10.1109/TPAMI.2010.57}
}

@inproceedings{babenko2014aq,
  title     = {Additive Quantization for Extreme Vector Compression},
  author    = {Babenko, Artem and Lempitsky, Victor},
  booktitle = {Proceedings of the IEEE Conference on Computer Vision and Pattern Recognition},
  pages     = {931--938},
  year      = {2014},
  url       = {https://openaccess.thecvf.com/content_cvpr_2014/html/Babenko_Additive_Quantization_for_2014_CVPR_paper.html}
}

@inproceedings{zhang2014cq,
  title     = {Composite Quantization for Approximate Nearest Neighbor Search},
  author    = {Zhang, Ting and Du, Chao and Wang, Jingdong},
  booktitle = {Proceedings of the 31st International Conference on Machine Learning},
  series    = {Proceedings of Machine Learning Research},
  volume    = {32},
  pages     = {838--846},
  year      = {2014},
  url       = {https://proceedings.mlr.press/v32/zhangd14.html}
}

@inproceedings{yu2024magvitv2,
  title     = {Language Model Beats Diffusion -- Tokenizer is Key to Visual Generation},
  author    = {Yu, Lijun and Lezama, Jos{\'e} and Gundavarapu, Nitesh B. and Versari, Luca and Sohn, Kihyuk and Minnen, David and Cheng, Yong and Birodkar, Vighnesh and Gupta, Agrim and Gu, Xiuye and Hauptmann, Alexander G. and Gong, Boqing and Yang, Ming-Hsuan and Essa, Irfan and Ross, David A. and Jiang, Lu},
  booktitle = {The Twelfth International Conference on Learning Representations},
  year      = {2024},
  url       = {https://openreview.net/forum?id=gzqrANCF4g}
}

@inproceedings{wang2024omnitokenizer,
  title     = {OmniTokenizer: A Joint Image-Video Tokenizer for Visual Generation},
  author    = {Wang, Junke and Jiang, Yi and Yuan, Zehuan and Peng, Binyue and Wu, Zuxuan and Jiang, Yu-Gang},
  booktitle = {Advances in Neural Information Processing Systems},
  volume    = {37},
  year      = {2024},
  url       = {https://proceedings.neurips.cc/paper_files/paper/2024/hash/31994923f58ae5b2d661b300bd439107-Abstract-Conference.html}
}

@inproceedings{zhao2025bsq,
  title     = {Image and Video Tokenization with Binary Spherical Quantization},
  author    = {Zhao, Yue and Xiong, Yuanjun and Kr{\"a}henb{\"u}hl, Philipp},
  booktitle = {The Thirteenth International Conference on Learning Representations},
  year      = {2025},
  url       = {https://openreview.net/forum?id=yGnsH3gQ6U}
}

@inproceedings{malepathirana2023napavq,
  title     = {{NAPA-VQ}: Neighborhood-Aware Prototype Augmentation with Vector Quantization for Continual Learning},
  author    = {Malepathirana, Tamasha and Senanayake, Damith and Halgamuge, Saman},
  booktitle = {Proceedings of the IEEE/CVF International Conference on Computer Vision},
  pages     = {11640--11650},
  year      = {2023},
  url       = {https://openaccess.thecvf.com/content/ICCV2023/html/Malepathirana_NAPA-VQ_Neighborhood-Aware_Prototype_Augmentation_with_Vector_Quantization_for_Continual_Learning_ICCV_2023_paper.html}
}

@inproceedings{jiao2024vqprompt,
  title     = {Vector Quantization Prompting for Continual Learning},
  author    = {Jiao, Li and Lai, Qiuxia and Li, Yu and Xu, Qiang},
  booktitle = {Advances in Neural Information Processing Systems},
  volume    = {37},
  year      = {2024},
  doi       = {10.52202/079017-1072},
  url       = {https://proceedings.neurips.cc/paper_files/paper/2024/hash/3baf4eeffad860ca9c54aeab632716b4-Abstract-Conference.html}
}

@inproceedings{ishikawa2021asrf,
  title     = {Alleviating Over-Segmentation Errors by Detecting Action Boundaries},
  author    = {Ishikawa, Yuchi and Kasai, Seito and Aoki, Yoshimitsu and Kataoka, Hirokatsu},
  booktitle = {Proceedings of the IEEE/CVF Winter Conference on Applications of Computer Vision (WACV)},
  pages     = {2322--2331},
  year      = {2021}
}

@inproceedings{lu2024fact,
  title     = {FACT: Frame-Action Cross-Attention Temporal Modeling for Efficient Action Segmentation},
  author    = {Lu, Zijia and Elhamifar, Ehsan},
  booktitle = {Proceedings of the IEEE/CVF Conference on Computer Vision and Pattern Recognition (CVPR)},
  pages     = {18175--18185},
  year      = {2024}
}

@inproceedings{rahaman2022timestamp,
  title     = {A Generalized and Robust Framework for Timestamp Supervision in Temporal Action Segmentation},
  author    = {Rahaman, Rahul and Singhania, Dipika and Thiery, Alexandre H. and Yao, Angela},
  booktitle = {Computer Vision -- ECCV 2022},
  series    = {Lecture Notes in Computer Science},
  volume    = {13664},
  pages     = {279--296},
  year      = {2022},
  doi       = {10.1007/978-3-031-19772-7_17}
}

@article{ding2022activitylabels,
  title   = {Temporal Action Segmentation With High-Level Complex Activity Labels},
  author  = {Ding, Guodong and Yao, Angela},
  journal = {IEEE Transactions on Multimedia},
  year    = {2022},
  doi     = {10.1109/TMM.2022.3231099}
}

@inproceedings{kumar2022unsupervised,
  title     = {Unsupervised Action Segmentation by Joint Representation Learning and Online Clustering},
  author    = {Kumar, Sateesh and Haresh, Sanjay and Ahmed, Awais and Konin, Andrey and Zia, M. Zeeshan and Tran, Quoc-Huy},
  booktitle = {Proceedings of the IEEE/CVF Conference on Computer Vision and Pattern Recognition (CVPR)},
  pages     = {20174--20185},
  year      = {2022}
}

@inproceedings{ding2024tca,
  title     = {Coherent Temporal Synthesis for Incremental Action Segmentation},
  author    = {Ding, Guodong and Golong, Hans and Yao, Angela},
  booktitle = {Proceedings of the IEEE/CVF Conference on Computer Vision and Pattern Recognition (CVPR)},
  pages     = {28485--28494},
  year      = {2024}
}

@inproceedings{ding2025gni,
  title     = {Condensing Action Segmentation Datasets via Generative Network Inversion},
  author    = {Ding, Guodong and Chen, Rongyu and Yao, Angela},
  booktitle = {Proceedings of the IEEE/CVF Conference on Computer Vision and Pattern Recognition (CVPR)},
  pages     = {17733--17742},
  year      = {2025}
}

@article{simeoni2025dinov3,
  title={Dinov3},
  author={Sim{\'e}oni, Oriane and Vo, Huy V and Seitzer, Maximilian and Baldassarre, Federico and Oquab, Maxime and Jose, Cijo and Khalidov, Vasil and Szafraniec, Marc and Yi, Seungeun and Ramamonjisoa, Micha{\"e}l and others},
  journal={arXiv preprint arXiv:2508.10104},
  year={2025}
}

@inproceedings{chae2025variable,
  title={Variable bitrate residual vector quantization for audio coding},
  author={Chae, Yunkee and Choi, Woosung and Takida, Yuhta and Koo, Junghyun and Ikemiya, Yukara and Zhong, Zhi and Cheuk, Kin Wai and Mart{\'\i}nez-Ram{\'\i}rez, Marco A and Lee, Kyogu and Liao, Wei-Hsiang and others},
  booktitle={ICASSP 2025-2025 IEEE International Conference on Acoustics, Speech and Signal Processing (ICASSP)},
  pages={1--5},
  year={2025},
  organization={IEEE}
}

\newif\ifshowappendix

\showappendixtrue

\ifshowappendix
    \newpage
    \appendix
    \section{Broader Impacts}\label{app:broader}
    ARQ-VAE can make long-horizon video understanding more practical by reducing the cost of storing and replaying dense frame features, especially in incremental and continual learning settings with limited memory. This may lower barriers for long-form video understanding, robotics, and embodied AI systems in resource-constrained environments. Although we believe our work is unlikely to have direct negative societal impacts, we acknowledge the importance of considering potential misuse scenarios, such as the exploitation of continually adapting systems in unethical surveillance or misinformation campaigns.

    \section{Storage Accounting}\label{app:storage}
    A central advantage of our representation is that its storage cost is explicit.
    After compression, the memory consists of stored residual indices and the active code vectors that remain after pruning unused entries.
    Let $L$ denote the total number of stored frames, and let $m_t$ be the number of retained indices for frame $t$.
    The index payload is
    \[
    S_{\mathrm{idx}} = \left(\sum_{t=1}^{L} m_t\right)\log_2 K \quad \text{bits},
    \]
    where $m_t=M$ for standard RQ-VAE and $m_t=m_t^*$ for ARQ-VAE.
    Since we use power-of-two codebook sizes, i.e., $K=2^x$, each index requires exactly $x=\log_2 K$ bits.

    Let $A_m$ be the number of active code vectors remaining in stage $m$ after pruning.
    Under float32 storage, the codebook cost is
    \[
    S_{\mathrm{code}} = 32d \sum_{m=1}^{M} A_m \quad \text{bits},
    \]
    where $d$ is the latent dimension.
    The total storage is therefore
    \begin{equation}
    S_{\mathrm{total}} = S_{\mathrm{idx}} + S_{\mathrm{code}}.
    \label{eq:storage_total}
    \end{equation}
    We report storage after pruning inactive codes.
    We do not include decoder parameters in the main storage tables~(per prior work~\cite{ding2025gni}), and we do not store a separate prefix-length token, since each frame is already represented by a truncated list of residual indices.

    \section{Comparison with Standard Feature-Compression Methods}
    \label{app:compression}

    We compare ARQ-VAE against commonly used feature-compression baselines: FP16 downcasting, deterministic per-channel INT8 quantization, lossless compression (\texttt{np.savez\_compressed}), and PCA with $d$ retained components. INT8 calibration is performed once during preprocessing; the evaluated I3D feature bank is nonnegative, with an observed range of approximately $[0, 5.18]$. Unlike the main tables, storage here is the complete footprint, including all method-specific parameters: PCA projections, INT8 metadata, and the decoder for RQ-VAE and ARQ-VAE. Results use MS-TCN on the same single 50Salads train/test split as Appendix~\ref{app:selection}, evaluated on the original, uncompressed test features. In this experiment, ARQ-VAE requires 3.152~MB under the paper's payload convention and 5.521~MB once the 2.369~MB decoder is included, compared with 3777.877~MB for the original FP32 training bank, so the compression remains substantial under either accounting convention.

    \begin{table}[h]
        \centering
        \small
        \setlength{\tabcolsep}{3.5pt}
        \caption{\textbf{Comparison with standard feature compression} on 50Salads with MS-TCN. Storage is the complete footprint, including all method-specific parameters (e.g., the decoder for RQ-VAE and ARQ-VAE).}
        \label{tab:compression}
        \resizebox{\linewidth}{!}{\begin{tabular}{lcccccccc}
            \toprule
            Method & Complete storage (MB) & Compression & MSE $\downarrow$ & MAE $\downarrow$ & Acc & Edit & F1@\{10,25,50\} \\
            \midrule
            Original FP32 & 3777.877 & 1.00$\times$ & 0 & 0 & 81.61 & 70.79 & 76.39 / 73.61 / 64.81 \\
            FP16 & 1888.947 & 2.00$\times$ & ${\approx}0$ & 0.00004 & 81.43 & 73.34 & 77.86 / 75.52 / 67.13 \\
            INT8 & 944.487 & 4.00$\times$ & 0.00002 & 0.00312 & 80.30 & 63.42 & 69.62 / 66.67 / 58.23 \\
            Lossless & 3331.176 & 1.13$\times$ & 0 & 0 & 81.61 & 70.79 & 76.39 / 73.61 / 64.81 \\
            \midrule
            PCA, $d{=}2$ & 3.725 & 1014.18$\times$ & 0.02948 & 0.11284 & 46.92 & 39.90 & 37.83 / 33.48 / 25.22 \\
            PCA, $d{=}4$ & 7.431 & 508.41$\times$ & 0.02732 & 0.10837 & 67.58 & 44.69 & 49.71 / 46.21 / 37.67 \\
            Fixed-8 RQ-VAE & 7.574 & 498.77$\times$ & 0.02605 & 0.10664 & 69.84 & 59.87 & 67.88 / 63.78 / 54.67 \\
            ARQ-VAE (L2 argmin) & 5.521 & 684.30$\times$ & 0.02068 & 0.09296 & 74.92 & 62.01 & 69.20 / 65.18 / 57.59 \\
            \bottomrule
        \end{tabular}}
    \end{table}

    \begin{figure}[h]
        \centering
        \includegraphics[width=\linewidth]{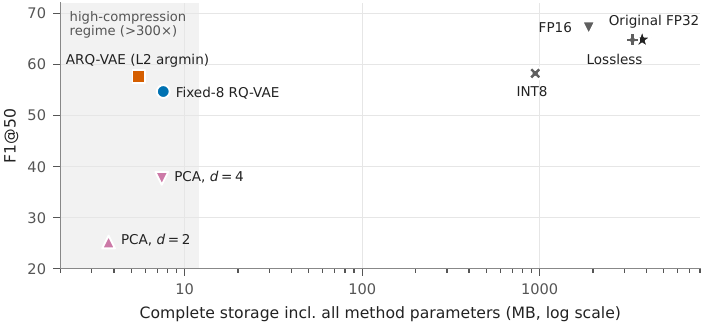}
        \caption{\textbf{Storage--performance trade-off against standard feature compression} on 50Salads with MS-TCN. Storage is the complete footprint, including all method-specific parameters, on a log scale. In the high-compression regime, ARQ-VAE retains far more downstream TAS performance than PCA at comparable storage.}
        \label{fig:compression}
    \end{figure}

    As shown in~\cref{tab:compression} and~\cref{fig:compression}, FP16, INT8, and lossless compression preserve feature quality but provide only modest storage reductions of $1.13\times$ to $4\times$. PCA reaches the same high-compression regime as our method, yet loses substantially more downstream utility: at roughly 7.5~MB, Fixed-8 RQ-VAE achieves 54.67 F1@50 versus 37.67 for PCA with $d{=}4$. ARQ-VAE improves this trade-off further, reducing complete storage to 5.521~MB while reaching 57.59 F1@50. ARQ-VAE is therefore most advantageous in the high-compression regime, where conventional precision reduction remains comparatively large and aggressive dimensionality reduction loses substantial downstream TAS utility.

    \section{Baseline Details}
    \label{app:baselines}

    This section gives additional implementation details for the baselines used in the supervised, incremental, and unsupervised TAS experiments. We follow the baseline definitions from TCA~\cite{ding2024tca} and GNI~\cite{ding2025gni} whenever applicable.

    \paragraph{Original features.}
    The \textbf{Original} baseline trains the downstream TAS model directly on the full pre-extracted feature bank. For a training video with frame features $X=\{x_t\}_{t=1}^{T}$, the model receives the original dense sequence $X$ and the corresponding frame-wise labels. No feature compression, condensation, replay synthesis, or reconstruction is applied. In incremental TAS, \textbf{Original} replay stores and reuses the original feature sequences from previous tasks, and therefore serves as an uncompressed replay reference under the same task protocol.

    \paragraph{Mean.}
    The \textbf{Mean} baseline compresses each ground-truth action segment into a single average feature vector. Let segment $s_n$ contain frame indices $\mathcal{I}_n=\{t_n,\dots,t_n+\ell_n-1\}$ with length $\ell_n$. Mean stores
    \[
    \bar{x}_n = \frac{1}{\ell_n}\sum_{t\in \mathcal{I}_n} x_t .
    \]
    During TAS training or replay, this stored vector is repeated $\ell_n$ times to restore the original temporal length of the segment. Thus, the reconstructed segment is static:
    \[
    \hat{x}_t = \bar{x}_n \quad \text{for all } t\in \mathcal{I}_n .
    \]
    This baseline primarily compresses the temporal dimension, reducing a video from frame-level storage to one feature vector per action segment.

    \paragraph{Coreset.}
    The \textbf{Coreset} baseline uses the same segment-level compression ratio as Mean, but stores a real frame feature instead of the averaged feature. For each segment $s_n$, we first compute the segment mean $\bar{x}_n$ as above and then select the frame feature nearest to it:
    \[
    x_n^{\mathrm{core}} = x_{t_n^*},
    \qquad
    t_n^* = \argmin_{t\in \mathcal{I}_n} \|x_t-\bar{x}_n\|_2^2 .
    \]
    During training or replay, $x_n^{\mathrm{core}}$ is repeated for all $\ell_n$ frames in the segment. The difference from Mean is therefore the stored representative: Mean stores the arithmetic average, while Coreset stores the closest actual frame feature.

    \paragraph{TCA.}
    \textbf{TCA}~\cite{ding2024tca} is a label-conditioned generative replay baseline for incremental TAS. It trains a VAE-style generator over frame features conditioned on the action label $a$ and a temporal coherence variable $c$, where $c$ encodes the relative position of a frame inside an action segment. For replay, a segment is generated by sampling a latent code $z$, fixing it across the segment, varying $c$ from the start to the end of the segment, and decoding features as
    \[
    \hat{x}_i = D_{\theta}(z,a,c_i), \qquad i=1,\dots,\ell .
    \]
    Generated segments are concatenated according to sampled action sequences and durations to form replay videos. In the TCA baseline, the latent codes are sampled from a normal prior during generation rather than stored as optimized per-segment representations.

    \paragraph{Encoded.}
    \textbf{Encoded} is the non-inversion comparison used in GNI~\cite{ding2025gni}. It uses the trained TCA encoder to map each frame in a segment to a latent distribution and stores the average encoded latent for that segment. If the encoder produces latent means $\mu_t$ for frames $t\in\mathcal{I}_n$, Encoded stores
    \[
    z_n^{\mathrm{enc}} = \frac{1}{\ell_n}\sum_{t\in\mathcal{I}_n} \mu_t .
    \]
    At reconstruction time, this latent is decoded with the TCA decoder using the segment action label and coherence variable. Unlike GNI, Encoded does not optimize the latent code by network inversion; it simply averages the encoder outputs.

    \paragraph{GNI.}
    \textbf{GNI}~\cite{ding2025gni} builds on the TCA generator but replaces averaged encoded latents with optimized segment latents. After training the TCA generator, GNI freezes the decoder and optimizes a latent code for each segment so that the decoded segment features match the original segment features:
    \[
    z_n^* = \argmin_z \sum_{i=1}^{\ell_n}
    \|D_{\theta}(z,a_n,c_i)-x_{t_n+i-1}\|_2^2 .
    \]
    The optimized latent codes are then stored as the condensed dataset and decoded during downstream TAS training or replay. Thus, GNI performs segment-level condensation in the latent space of a label-conditioned generator.

    \paragraph{Finetune.}
    \textbf{Finetune} is used only in the incremental TAS setting. At each task step, the TAS model is trained on the current task data only, using the original current-task features and labels. No previous-task data, replay features, stored exemplars, or generated samples are used. This baseline measures catastrophic forgetting when the model adapts to new tasks without any memory mechanism.

    \paragraph{Exemplar.}
    \textbf{Exemplar} is the direct-replay baseline used in TCA~\cite{ding2024tca}. For each previous-task action segment, it stores one representative real frame feature. The stored frame is then repeated across the segment duration to form a static replay segment, analogous to Coreset-style temporal inflation:
    \[
    \hat{x}_t = x_n^{\mathrm{ex}} \quad \text{for all } t\in\mathcal{I}_n .
    \]
    This baseline stores actual features rather than generated or quantized features, but it loses temporal variation within each action segment because replay segments are static after repetition.

    \paragraph{VQ-VAE.}
    \textbf{VQ-VAE} is our single-stage vector-quantized autoencoder baseline. It uses one discrete codebook to quantize each frame feature into a single code index and reconstructs the feature through the decoder. This is equivalent to using one quantization stage, i.e., $M=1$, and therefore tests whether a single discrete code per frame is sufficient for compact TAS feature memory.

    \paragraph{RQ-VAE.}
    \textbf{RQ-VAE} is the fixed-rate residual quantization baseline. Each frame feature is encoded into a sequence of $M$ residual code indices, and every frame stores all $M$ indices. Unlike ARQ-VAE, RQ-VAE does not adapt the number of retained residual stages per frame. It therefore isolates the benefit of residual quantization itself from the benefit of adaptive prefix truncation.

    \paragraph{ASOT setting for unsupervised TAS.}
    For unsupervised TAS, we use ASOT~\cite{xu2024ASOT} as the downstream segmentation method. The \textbf{Original} unsupervised baseline trains ASOT with the original I3D feature bank. Our RQ-VAE, ARQ-VAE, and ARQ-VAE$^*$ rows replace only the training feature bank with reconstructed features, while keeping the ASOT procedure unchanged. This evaluates whether the compressed representation remains useful without relying on frame-level action labels during downstream training.

    \section{Additional Implementation Details}
    \label{app:implementation}

    Unless otherwise stated, we use $K{=}128$ codes per stage and $M{=}8$ residual stages for all main ARQ-VAE experiments. The reconstruction summary in \cref{tab:oracle-summary} averages over a broader grid of $(K,M)$ settings, but we use this single default configuration for downstream TAS experiments; empirically, performance is not highly sensitive to the exact choice of $K$ and $M$ after adaptive prefix selection. We train RQ-VAE and ARQ-VAE with Adam using a batch size of 256, learning rate $10^{-3}$, and commitment weight $\beta{=}0.25$. I3D features are used directly without additional normalization. Mini-batches for ARQ-VAE training are sampled uniformly over all training frames.

    For downstream TAS training, we store the adaptive code representation and reconstruct features on the fly. The dataloader loads the retained indices for each frame, retrieves the corresponding active codebook entries, and applies the decoder during collation to produce reconstructed training features. TAS backbones are still trained video-wise, following the standard MS-TCN and ASFormer setup, rather than with uniformly sampled frame batches. For storage accounting, we count the retained discrete indices and the pruned active codebooks. The encoder is discarded, unused codebook entries are removed, and prefix lengths are not stored separately because each frame is represented directly by its retained variable-length index sequence. Supervised TAS results use full cross-validation over the official dataset splits: four splits for Breakfast and GTEA, and five splits for 50Salads. For incremental TAS, we use a single split for Breakfast and YouTube Instructional and report averages over three unique task sequences. For supervised and unsupervised experiments, ARQ-VAE is trained at the dataset-split level on the full training feature bank. For incremental TAS, ARQ-VAE is instead trained on each task subset sequentially, matching the iTAS protocol: after a task is compressed, only its compact ARQ-VAE memory is retained for replay in later tasks.

    \section{Ablation on Number of Codes}
    \label{app:width-ablation}

    This ablation studies the effect of codebook width $K$ while keeping the number of residual stages fixed at $M{=}8$. The goal is to evaluate how the size of each stage-level codebook affects the storage--performance tradeoff. As shown in Table~\ref{tab:width_ablation}, increasing $K$ generally improves reconstruction capacity but also increases storage, and fixed-depth RQ-VAE does not always translate larger codebooks into better TAS performance. ARQ-VAE variants are more robust across widths because they retain only the useful residual prefixes, with ARQ-VAE$^*$ giving the strongest overall balance between task performance, reconstruction quality, and compactness.

    \begin{table*}[t]
        \centering
        \small
        \setlength{\tabcolsep}{4.5pt}
        \caption{Width ablation with fixed depth $M{=}8$. Higher is better for task metrics; lower is better for reconstruction metrics. Average and standard deviation are computed across the six tested width settings for each method.}
        \label{tab:width_ablation}
        \begin{adjustbox}{max width=\textwidth}
            \begin{tabular}{llccccccccc}
                \toprule
                \multirow{2}{*}{Method} & \multirow{2}{*}{$K$} & \multicolumn{6}{c}{Task metrics} & \multicolumn{3}{c}{Reconstruction metrics} \\
                \cmidrule(lr){3-8} \cmidrule(lr){9-11}
                 &  & Acc & Edit & F1@10 & F1@25 & F1@50 & AVG & MAE & Rel L2 & Storage (MB) \\
                \midrule
                \multirow{8}{*}{RQ-VAE}
                & 32   & 70.5 & 56.2 & 63.2 & 59.1 & 48.8 & 59.6 & 0.104 & 0.512 & \textbf{2.433} \\
                & 64   & 68.3 & 55.4 & 61.7 & 58.0 & 48.5 & 58.4 & 0.107 & 0.516 & 3.144 \\
                & 128  & 72.2 & 58.6 & 64.6 & 61.5 & 52.3 & 61.8 & \textbf{0.106} & \textbf{0.510} & 4.100 \\
                & 256  & 70.5 & 58.6 & 64.8 & 62.0 & 52.0 & 61.6 & 0.107 & \textbf{0.510} & 5.445 \\
                & 512  & \textbf{73.3} & 60.4 & \textbf{66.5} & \textbf{63.0} & \textbf{54.2} & \textbf{63.5} & 0.122 & 0.570 & 7.199 \\
                & 1024 & 70.6 & \textbf{61.3} & 66.4 & \textbf{63.0} & 52.8 & 62.8 & 0.130 & 0.610 & 12.763 \\
                \cmidrule(lr){2-11}
                \rowcolor{gray!8}
                & Average   & 70.9 & 58.4 & 64.5 & 61.1 & 51.4 & 61.3 & 0.113 & 0.538 & 5.847 \\
                \rowcolor{gray!8}
                & Std. Dev. & 1.56 & 2.09 & 1.69 & 1.91 & 2.09 & 1.77 & 0.010 & 0.039 & 3.460 \\
                \midrule
                \multirow{8}{*}{ARQ-VAE}
                & 32   & 72.0 & 55.1 & 60.4 & 56.1 & 47.7 & 58.3 & 0.098 & 0.485 & \textbf{1.374} \\
                & 64   & 74.3 & \textbf{63.5} & \textbf{69.7} & \textbf{66.5} & \textbf{57.2} & \textbf{66.2} & 0.095 & 0.469 & 1.743 \\
                & 128  & 73.9 & 60.6 & 64.7 & 61.6 & 52.4 & 62.6 & 0.093 & 0.458 & 2.400 \\
                & 256  & \textbf{75.0} & 62.6 & 69.0 & 66.4 & 56.4 & 65.9 & 0.092 & 0.449 & 3.464 \\
                & 512  & 74.7 & 60.2 & 66.7 & 63.3 & 54.7 & 63.9 & \textbf{0.091} & \textbf{0.444} & 4.388 \\
                & 1024 & 75.0 & 62.1 & 67.9 & 63.7 & 54.7 & 64.7 & \textbf{0.091} & 0.445 & 5.809 \\
                \cmidrule(lr){2-11}
                \rowcolor{gray!8}
                & Average   & 74.2 & 60.7 & 66.4 & 62.9 & 53.9 & 63.6 & 0.093 & 0.458 & 3.196 \\
                \rowcolor{gray!8}
                & Std. Dev. & 1.04 & 2.74 & 3.13 & 3.51 & 3.14 & 2.66 & 0.002 & 0.015 & 1.548 \\
                \midrule
                \multirow{8}{*}{ARQ-VAE$^*$}
                & 32   & 74.3 & 57.4 & 61.5 & 57.7 & 49.2 & 60.0 & 0.095 & 0.464 & 1.374 \\
                & 64   & 74.2 & \textbf{65.2} & \textbf{70.4} & \textbf{67.0} & 57.4 & \textbf{66.8} & 0.091 & 0.444 & 1.743 \\
                & 128  & 75.3 & 61.7 & 68.6 & 65.3 & 56.8 & 65.5 & 0.089 & 0.432 & 2.400 \\
                & 256  & 76.0 & 62.5 & 69.3 & 65.6 & 57.2 & 66.1 & 0.086 & 0.420 & 3.464 \\
                & 512  & \textbf{76.4} & 63.6 & 69.7 & 65.7 & \textbf{57.9} & 66.7 & \textbf{0.085} & \textbf{0.413} & 4.388 \\
                & 1024 & 74.7 & 64.1 & 69.3 & 65.8 & 56.6 & 66.1 & \textbf{0.085} & 0.414 & 5.809 \\
                \cmidrule(lr){2-11}
                \rowcolor{gray!8}
                & Average   & 75.1 & 62.4 & 68.1 & 64.5 & 55.9 & 65.2 & 0.089 & 0.431 & 3.196 \\
                \rowcolor{gray!8}
                & Std. Dev. & 0.83 & 2.51 & 3.01 & 3.09 & 3.00 & 2.37 & 0.004 & 0.018 & 1.548 \\
                \bottomrule
            \end{tabular}
        \end{adjustbox}
    \end{table*}

    \section{Ablation on Number of Residual Stages}
    \label{app:depth-ablation}

    This ablation studies the effect of residual depth $M$ while keeping the codebook width fixed at $K{=}128$. The goal is to test whether adding more residual stages consistently improves reconstruction and downstream TAS performance, or whether fixed-depth quantization becomes brittle when too many stages are forced onto every frame. As shown in Table~\ref{tab:depth_ablation}, fixed-depth RQ-VAE is sensitive to $M$: moderate depths work best, while very large depth can hurt both reconstruction and task performance. ARQ-VAE is more stable because it can discard unnecessary later stages on a per-frame basis, and ARQ-VAE$^*$ performs best overall after adapting the decoder and active codebooks to the retained prefixes.

    \begin{table*}[t]
        \centering
        \small
        \setlength{\tabcolsep}{4.5pt}
        \caption{Depth ablation with fixed width $K{=}128$. Higher is better for task metrics; lower is better for reconstruction metrics. Average and standard deviation are computed across the five tested depth settings for each method.}
        \label{tab:depth_ablation}
        \begin{adjustbox}{max width=\textwidth}
            \begin{tabular}{llccccccccc}
                \toprule
                \multirow{2}{*}{Method} & \multirow{2}{*}{$M$} & \multicolumn{6}{c}{TAS metrics} & \multicolumn{3}{c}{Reconstruction metrics} \\
                \cmidrule(lr){3-8} \cmidrule(lr){9-11}
                 &  & Acc & Edit & F1@10 & F1@25 & F1@50 & AVG & MAE & Rel L2 & Storage (MB) \\
                \midrule
                \multirow{7}{*}{RQ-VAE}
                & 1  & 69.6 & 56.7 & 60.9 & 57.8 & 48.8 & 58.8 & 0.103 & 0.506 & \textbf{0.467} \\
                & 2  & \textbf{73.9} & 60.1 & 65.1 & 61.5 & 53.6 & 62.8 & \textbf{0.097} & 0.478 & 1.027 \\
                & 4  & 73.2 & \textbf{63.5} & \textbf{69.8} & \textbf{65.7} & \textbf{56.6} & \textbf{65.8} & 0.098 & \textbf{0.476} & 2.053 \\
                & 8  & 72.2 & 58.6 & 64.6 & 61.5 & 52.3 & 61.8 & 0.106 & 0.510 & 4.100 \\
                & 16 & 63.5 & 50.7 & 55.3 & 51.5 & 41.7 & 52.5 & 0.202 & 0.926 & 7.349 \\
                \cmidrule(lr){2-11}
                \rowcolor{gray!8}
                & Average   & 70.5 & 57.9 & 63.1 & 59.6 & 50.6 & 60.3 & 0.121 & 0.579 & 2.999 \\
                \rowcolor{gray!8}
                & Std. Dev. & 3.82 & 4.24 & 4.84 & 4.84 & 5.08 & 4.54 & 0.041 & 0.174 & 2.503 \\
                \midrule
                \multirow{7}{*}{ARQ-VAE}
                & 1  & 69.6 & 56.7 & 60.9 & 57.8 & 48.8 & 58.8 & 0.103 & 0.506 & \textbf{0.467} \\
                & 2  & 73.0 & 56.9 & 61.6 & 58.2 & 50.1 & 60.0 & 0.096 & 0.470 & 0.914 \\
                & 4  & \textbf{73.9} & \textbf{62.1} & \textbf{66.8} & \textbf{63.2} & \textbf{53.8} & \textbf{64.0} & \textbf{0.093} & \textbf{0.456} & 1.556 \\
                & 8  & 73.9 & 60.6 & 64.7 & 61.6 & 52.4 & 62.6 & \textbf{0.093} & 0.458 & 2.400 \\
                & 16 & 70.1 & 55.4 & 59.8 & 55.4 & 47.5 & 57.6 & 0.096 & 0.478 & 1.881 \\
                \cmidrule(lr){2-11}
                \rowcolor{gray!8}
                & Average   & 72.1 & 58.3 & 62.8 & 59.2 & 50.5 & 60.6 & 0.096 & 0.474 & 1.444 \\
                \rowcolor{gray!8}
                & Std. Dev. & 1.88 & 2.58 & 2.57 & 2.81 & 2.31 & 2.42 & 0.004 & 0.018 & 0.686 \\
                \midrule
                \multirow{7}{*}{ARQ-VAE$^*$}
                & 1  & 69.6 & 56.7 & 60.9 & 57.8 & 48.8 & 58.8 & 0.103 & 0.506 & 0.467 \\
                & 2  & 73.6 & 59.5 & 64.7 & 61.5 & 53.2 & 62.5 & 0.094 & 0.458 & 0.914 \\
                & 4  & \textbf{75.4} & \textbf{66.8} & \textbf{71.3} & \textbf{68.3} & \textbf{57.4} & \textbf{67.8} & 0.090 & 0.437 & 1.556 \\
                & 8  & 75.3 & 61.7 & 68.6 & 65.3 & 56.8 & 65.5 & \textbf{0.089} & \textbf{0.432} & 2.400 \\
                & 16 & 73.9 & 61.7 & 67.2 & 63.9 & 55.0 & 64.3 & 0.092 & 0.452 & 1.881 \\
                \cmidrule(lr){2-11}
                \rowcolor{gray!8}
                & Average   & 73.6 & 61.3 & 66.5 & 63.4 & 54.2 & 63.8 & 0.094 & 0.457 & 1.444 \\
                \rowcolor{gray!8}
                & Std. Dev. & 2.08 & 3.28 & 3.48 & 3.46 & 3.10 & 3.04 & 0.005 & 0.026 & 0.686 \\
                \bottomrule
            \end{tabular}
        \end{adjustbox}
    \end{table*}

    \section{Storage--Performance Plots for Width/Depth Ablations}
    \label{app:rate-distortion}

    \begin{figure*}[t]
        \centering
        \begin{subfigure}[t]{0.48\textwidth}
            \centering
            \includegraphics[width=\linewidth]{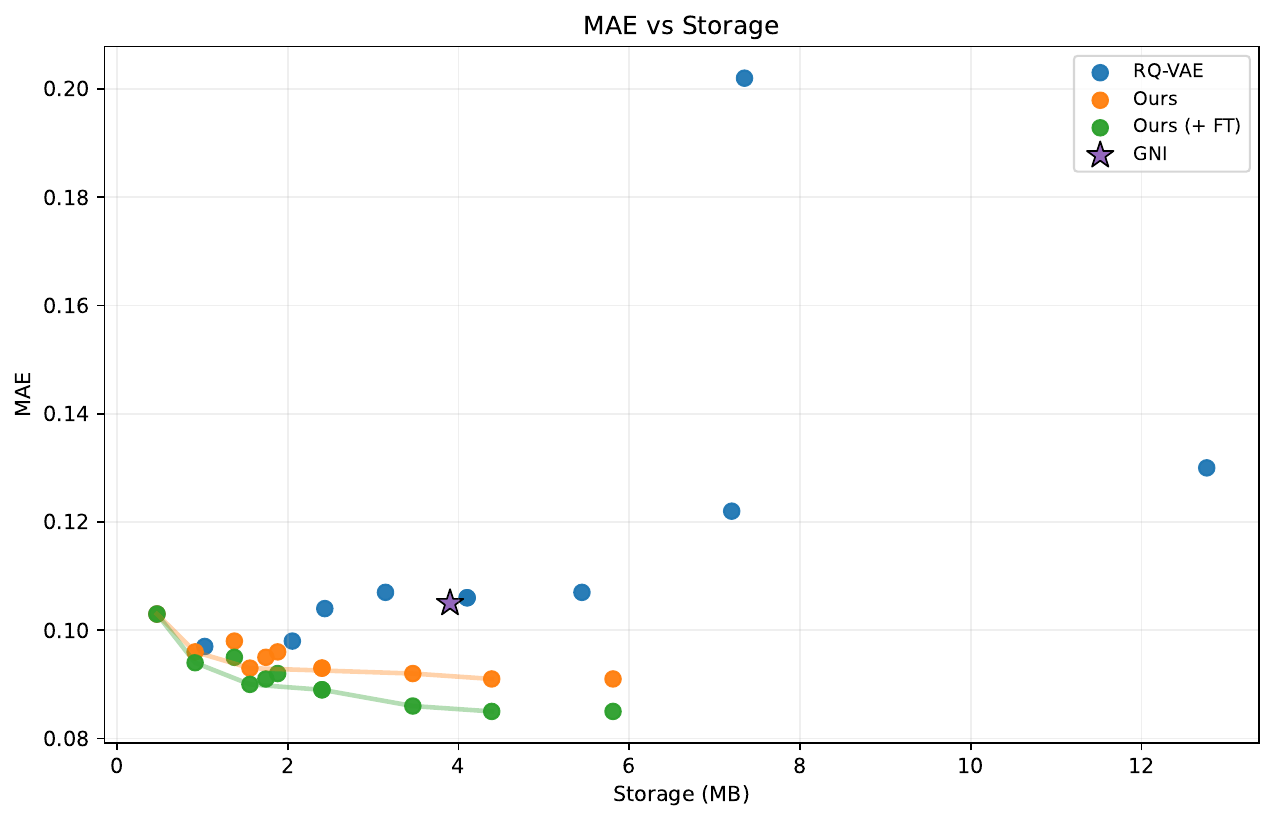}
            \caption{Reconstruction error (MAE) versus storage.}
            \label{fig:rd_mae}
        \end{subfigure}
        \hfill
        \begin{subfigure}[t]{0.48\textwidth}
            \centering
            \includegraphics[width=\linewidth]{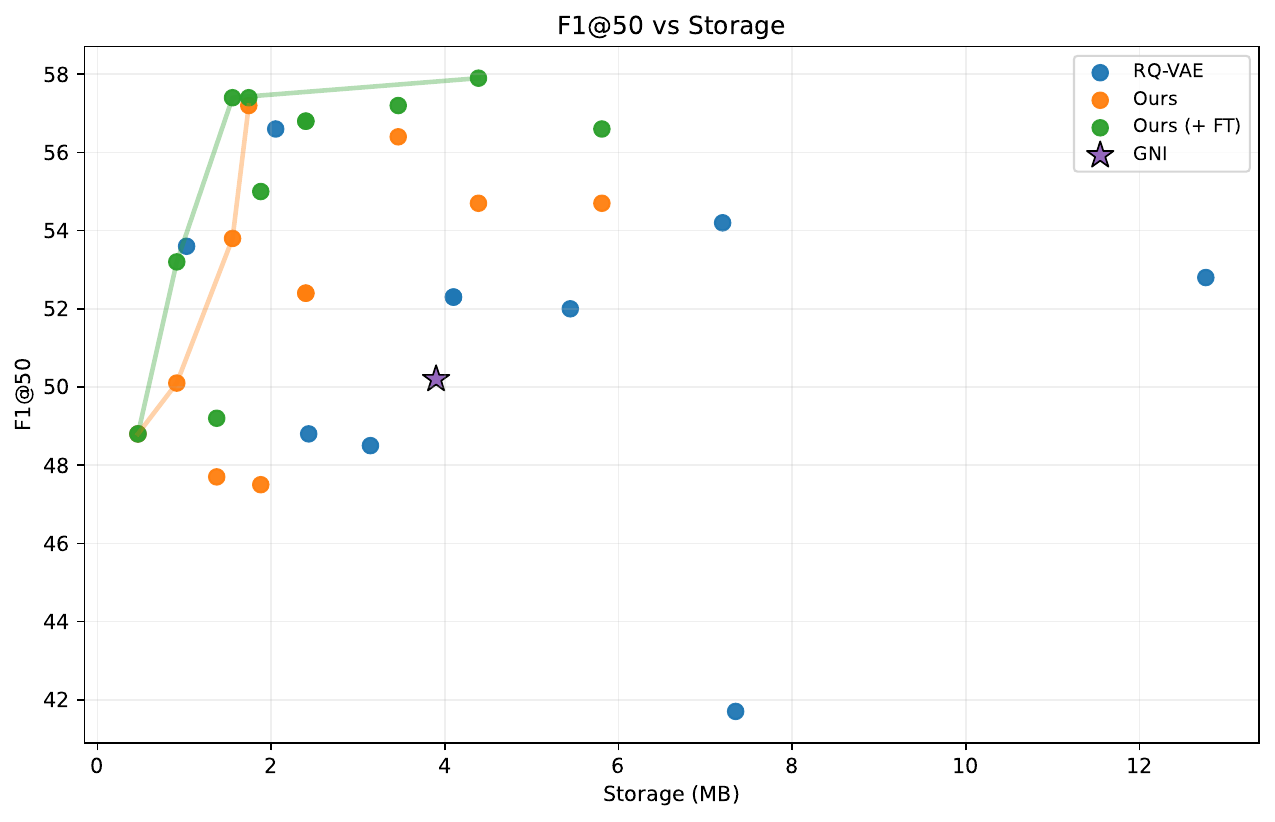}
            \caption{TAS performance (F1@50) versus storage.}
            \label{fig:rd_f1}
        \end{subfigure}
        \caption{Storage--performance plots for the width and depth ablations, using reconstruction mean absolute error and downstream temporal action segmentation performance.}
        \label{fig:rate_distortion}
    \end{figure*}

    \section{Prefix Length Histograms}\label{app:histo}

    \begin{figure*}[t]
        \centering
        \includegraphics[width=\linewidth]{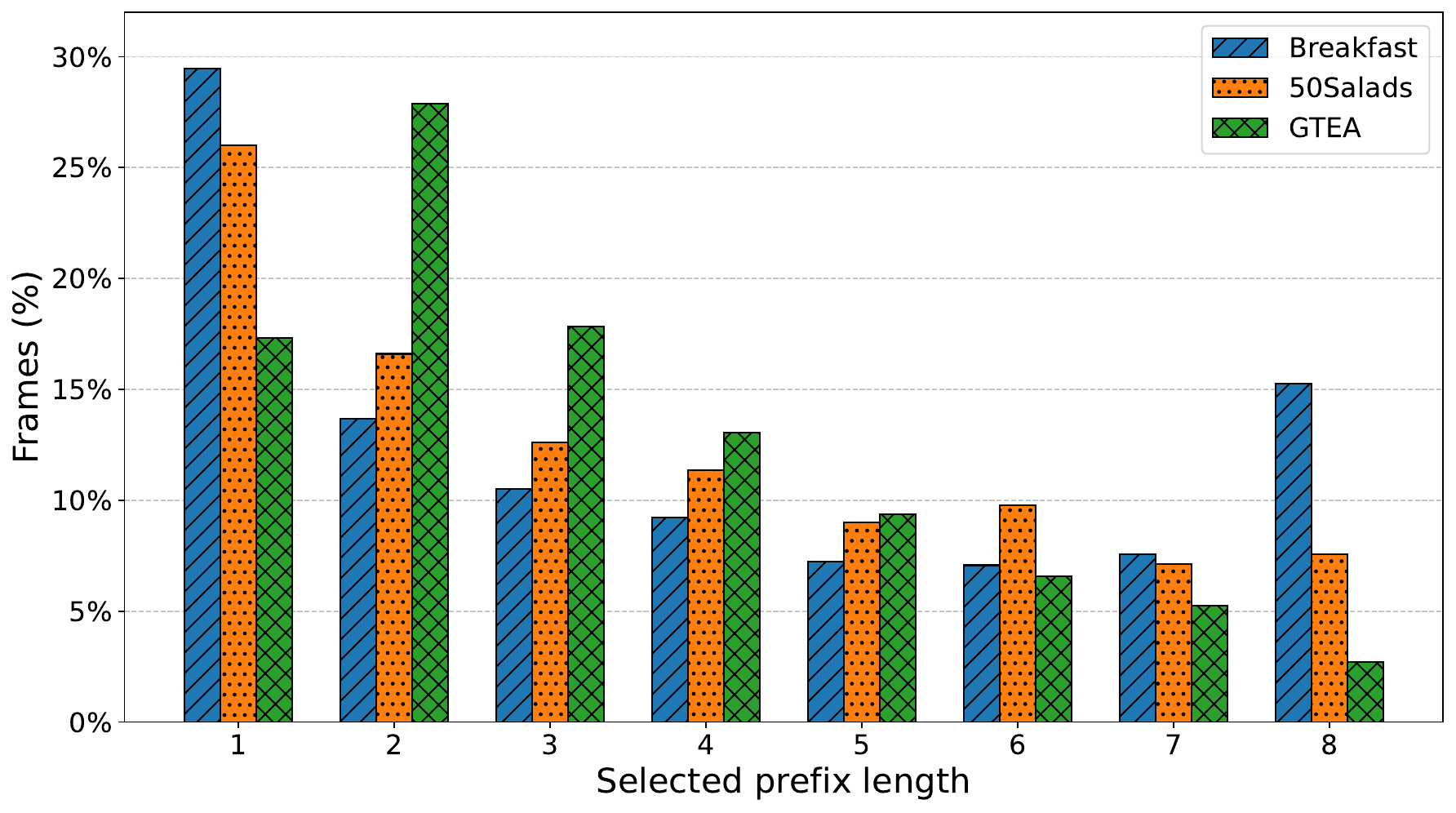}
        \caption{\textbf{Distribution of selected prefix lengths} across GTEA, 50Salads, and Breakfast.}
        \label{fig:prefix-histogram}
    \end{figure*}

    \Cref{fig:prefix-histogram} shows the distribution of selected ARQ-VAE prefix lengths across the three TAS benchmark datasets. The selected depths are spread across the full range of residual stages rather than collapsing to a single stage, indicating that frames vary substantially in how much residual capacity they need. Many frames are reconstructed with short prefixes, reflecting redundancy in stable action regions, while a nontrivial fraction retain deeper prefixes, corresponding to frames that benefit from additional refinement. This supports the central motivation for adaptive residual quantization: a fixed residual budget is unnecessarily rigid for long procedural videos, where different frames require different amounts of memory.

    \section{Prefix Length Tracks Reconstruction Difficulty} \label{app:prefix_difficulty}

    To validate the hypothesis that \emph{hard} reconstruction tasks require deeper residual refinement while \emph{easy} and temporally redundant reconstruction tasks can be represented with shorter prefixes, we analyze the frame-wise prefix lengths selected by ARQ-VAE after prefix selection. Recall that ARQ-VAE stores, for each training frame, the residual prefix that achieves the best reconstruction among all decoded prefixes. Under this view, the selected prefix length $m_t^\star$ serves as an operational measure of frame difficulty: frames that are already well reconstructed by shallow prefixes are assigned shorter codes, while frames that continue to benefit from additional residual refinement retain deeper prefixes.

    \paragraph{Protocol.}
    This analysis is performed on the \emph{training} feature bank, since ARQ-VAE compresses the stored training set and does not reconstruct test features. For this analysis, we globally normalize the pre-extracted frame features using the training-set mean and standard deviation. We then compute two frame-wise redundancy measures. \textbf{Pred.} measures \emph{local-context predictability}: a frame is more redundant if it is well explained by the average of its temporal neighbors. \textbf{Var.} measures \emph{local temporal variance}: a frame is more redundant if its feature vector lies in a temporally stable window with low variation. To capture how prefix length varies with redundancy over time, we compute Pearson and Spearman correlations between the raw prefix length $m_t^\star$ and each redundancy score within each video. For clarity, any moving-average prefix curve is used only for visualization and is \emph{not} used in the reported statistics. We report mean per-video Pearson and Spearman correlations (Mean P and Mean S), the fraction of videos with negative correlation (\% neg.\ videos), and pooled Pearson correlation across all frames (Pooled P) in Table~\ref{tab:prefix_redundancy}.

    \begin{table}[t]
    \centering
    \footnotesize
    \caption{\textbf{Frame-wise prefix length is inversely correlated with feature redundancy.}
    \emph{Pred.}\ denotes local-context predictability and \emph{Var.}\ denotes local temporal variance. Negative correlation means that more redundant frames tend to use shorter prefixes, and larger redundancy scores indicate greater temporal redundancy.} \vspace{.5em}
    \setlength{\tabcolsep}{10pt}
    \resizebox{\linewidth}{!}{\begin{tabular}{llcccc}
    \toprule
    Dataset & Metric & Mean P$\downarrow$ & Mean S$\downarrow$ & \% neg.\ vids$\uparrow$ & Pooled P$\downarrow$ \\
    \midrule
    \multirow{2}{*}{50Salads}  & Pred. & -0.2057 & -0.1966 & 100.0 & -0.2163 \\
              & Var.  & -0.1828 & -0.1847 & 97.5  & -0.1951 \\
    \midrule
    \multirow{2}{*}{GTEA}      & Pred. & -0.2856 & -0.2675 & 100.0 & -0.2999 \\
                  & Var.  & -0.2688 & -0.2524 & 100.0 & -0.2860 \\
    \midrule
    \multirow{2}{*}{Breakfast} & Pred. & -0.2135 & -0.2031 & 90.9  & -0.2587 \\
             & Var.  & -0.2046 & -0.1913 & 88.0  & -0.2511 \\
    \bottomrule
    \end{tabular}}
    \label{tab:prefix_redundancy}
    \end{table}
    \begin{figure}[t]
        \centering
        \caption{\textbf{Example video-level relationship between prefix length and feature redundancy.}
        Top: normalized trajectories of the selected prefix length and the feature redundancy score over time, with Pearson and Spearman correlations shown in the inset. Bottom: ground-truth action segmentation for the same video. Regions with sustained actions such as \texttt{spoon\_powder}, \texttt{pour\_milk}, and \texttt{stir\_milk} tend to exhibit higher redundancy and shorter selected prefixes, while lower-redundancy portions of the sequence coincide with longer prefixes. This example illustrates the inverse relationship quantified in Table~\ref{tab:prefix_redundancy}: ARQ-VAE allocates more residual depth to temporally harder, less redundant frames and fewer stages to stable action interiors.}
        \includegraphics[width=\linewidth]{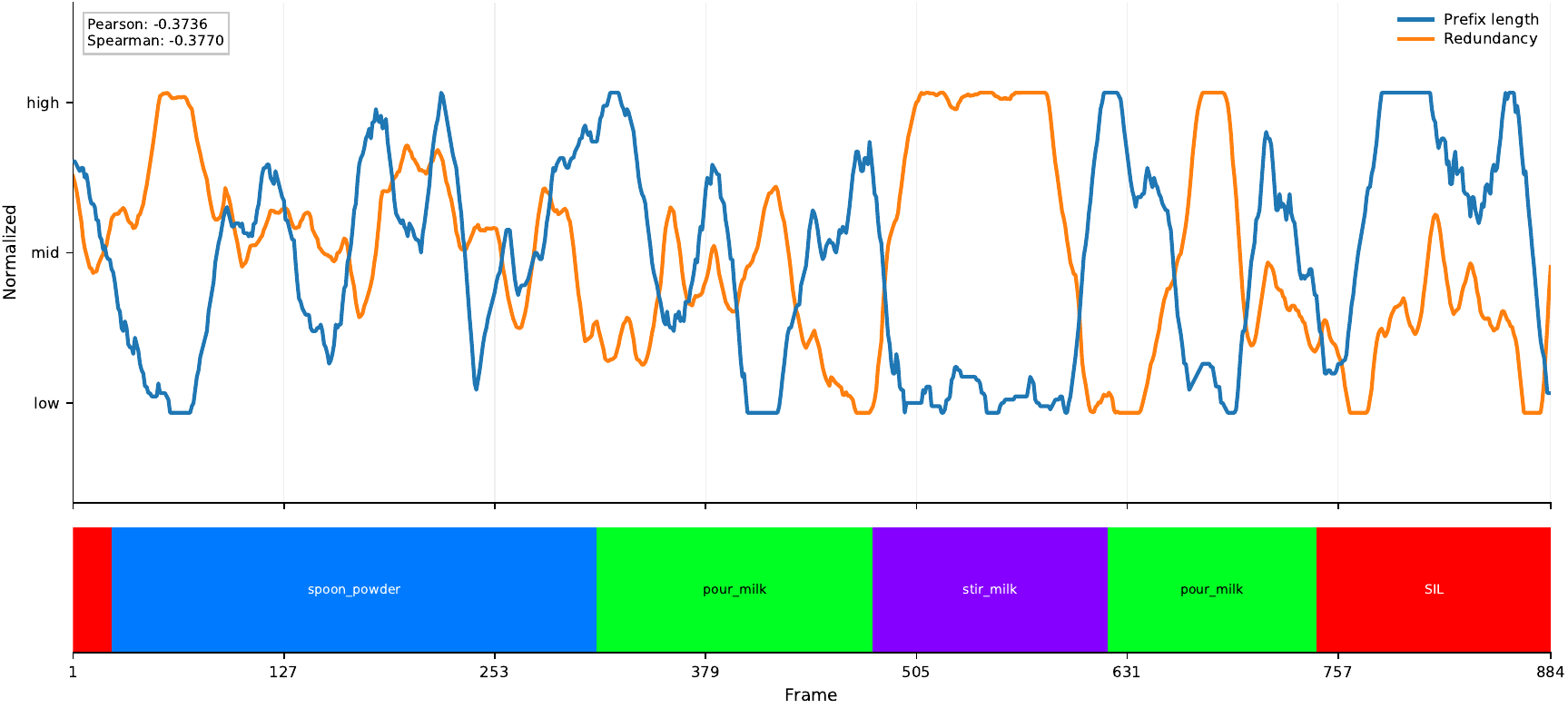}
        \label{fig:prefix_redundancy_example}
    \end{figure}

    \paragraph{Results.}
    Table~\ref{tab:prefix_redundancy} shows a consistent inverse relationship between redundancy and selected prefix length across all three datasets. The mean per-video Pearson and Spearman correlations are negative in every case. The sign is also highly stable across videos: for both metrics, the fraction of videos with negative correlation is at least 88.0\% on Breakfast and reaches 100\% on GTEA. The pooled frame-wise correlations are likewise uniformly negative. Thus, regardless of whether redundancy is measured by temporal predictability or local feature stability, frames that are easier to explain in context tend to terminate after fewer residual stages in ARQ-VAE.

    \paragraph{Interpretation.}
    These results provide direct evidence for the intended ARQ-VAE behavior. Prefix length is not assigned using labels, boundaries, or any external notion of frame difficulty; it emerges purely from the reconstruction-based prefix-selection rule. The negative correlations in Table~\ref{tab:prefix_redundancy}, together with the qualitative timeline visualization in Fig.~\ref{fig:prefix_redundancy_example}, show that ARQ-VAE allocates longer prefixes to rapidly changing or harder-to-reconstruct regions and shorter prefixes to stable, redundant regions. This supports the central narrative of the method: adaptive residual quantization spends capacity where the feature sequence is harder to preserve.

    \section{Qualitative Analysis of Reused Residual Codes}
    \label{app:code-semantics}

    \begin{figure*}[t]
        \centering
        \includegraphics[width=\linewidth]{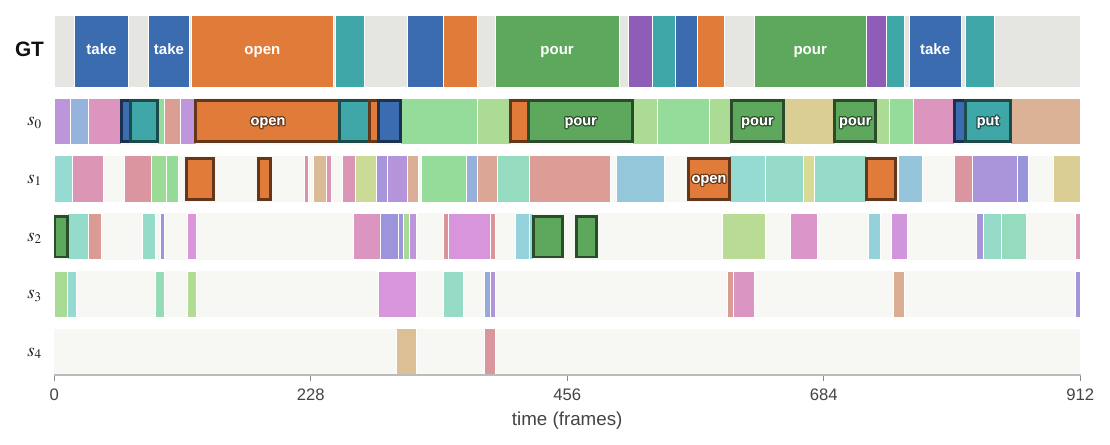}
        \caption{\textbf{Residual code reuse on a GTEA example.} The top row shows ground-truth action segments, while lower rows show frequently reused code indices at residual stages $s_0,\ldots,s_4$. Colored intervals indicate retained codes, and greyed-out regions indicate later-stage codes removed by ARQ-VAE prefix truncation. Reused codes often align with repeated actions such as \emph{take}, \emph{open}, and \emph{pour}, suggesting a weak correspondence between discrete codes and recurring procedural structure.}
        \label{fig:code-reuse-cheese}
    \end{figure*}

    \begin{figure*}[t]
        \centering
        \includegraphics[width=\linewidth]{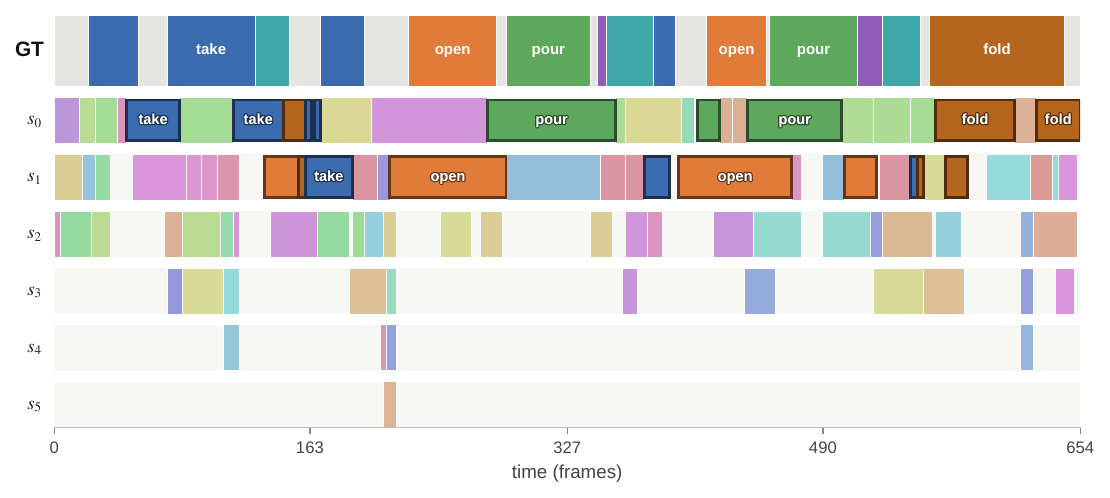}
        \caption{\textbf{Residual code reuse on a second GTEA example.} As in \Cref{fig:code-reuse-cheese}, early residual stages contain longer reused code intervals that overlap with major action segments, while later stages are more fragmented and are often discarded by adaptive prefix selection. Repeated codes appear across multiple instances of actions such as \emph{take}, \emph{open}, \emph{pour}, and \emph{fold}.}
        \label{fig:code-reuse-hotdog}
    \end{figure*}

    \Cref{fig:code-reuse-cheese,fig:code-reuse-hotdog} provide a qualitative view of how residual codes are reused over time. ARQ-VAE is trained only to reconstruct pre-extracted features, without action labels, but residual quantization still learns a shared discrete vocabulary. In procedural videos, repeated actions often have similar visual-temporal structure, so the same code indices can reappear within semantically related segments.

    Each figure shows ground-truth actions on top and reused code indices across residual stages below. The alignment is not one-to-one: a code is not an action label, and an action may use multiple codes depending on context, viewpoint, and temporal phase. Instead, the plots show a weaker pattern: early-stage codes often form coherent intervals that overlap with actions such as \emph{take}, \emph{open}, \emph{pour}, \emph{put}, and \emph{fold}. This suggests that the discrete memory captures recurring feature patterns rather than simply memorizing frame positions.

    The figures also visualize adaptive prefix selection. Colored blocks are retained codes, while greyed-out later stages are removed by ARQ-VAE. Early stages tend to preserve coarse action-level structure, whereas later stages are more fragmented and are kept mainly for frames needing extra reconstruction detail. This supports the intuition behind ARQ-VAE: common procedural patterns can reuse compact codes, while harder frames receive deeper residual prefixes only when needed.

    \section{\texorpdfstring{Prefix-Selection Criterion and Rate--Distortion Control}{Prefix-Selection Criterion and Rate-Distortion Control}}
    \label{app:selection}

    This section expands on~\cref{sec:selection}. All criteria use the same pretrained $M{=}8$, $K{=}128$ RQ-VAE; none retrains the base codec. These diagnostic experiments use a single standard train/test partition of 50Salads with the 50-epoch MS-TCN protocol, and TAS models are evaluated on the original, uncompressed test features. Storage follows the paper's convention (indices and active codebooks, decoder excluded), while Complete additionally includes the 2.369-MB decoder. \Cref{fig:selection} in the main text summarizes all criteria.

    \paragraph{Selection criteria.}
    \textbf{Fixed-8} retains all eight stages and is equal to RQ-VAE. \textbf{L2 argmin} selects the reconstruction-optimal prefix independently for each frame,~\cref{eq:oracle_prefix_main}, and is equal to ARQ-VAE. \textbf{L2-permuted} randomly reassigns the L2-selected depths across frames, preserving the exact depth histogram and storage of ARQ-VAE while removing the frame/depth association. The \textbf{label-aware probe} trains a linear frame classifier on the original training features and TAS labels, then selects the prefix with the minimum correct-class cross-entropy; it tests whether local label discriminability provides a better framewise signal than reconstruction error. We use a framewise probe because a standard TAS model such as MS-TCN is temporal: a forward pass requires the full frame sequence, so exactly optimizing per-frame prefix lengths for a segmentation loss could require up to $8^{T}$ sequence configurations and is computationally infeasible. The probe is trained on all labeled training frames, including frames near action transitions, but it does not explicitly model temporal ordering, boundaries, or segment-level metrics.

    \begin{table}[h]
        \centering
        \small
        \setlength{\tabcolsep}{3.5pt}
        \caption{\textbf{Prefix-selection criteria} on 50Salads with MS-TCN. Storage excludes the decoder; Complete includes it.}
        \label{tab:selection-criteria}
        \resizebox{\linewidth}{!}{\begin{tabular}{lcccccccc}
            \toprule
            Criterion & Mean depth & Storage (MB) & Complete (MB) & MSE $\downarrow$ & MAE $\downarrow$ & Acc & Edit & F1@\{10,25,50\} \\
            \midrule
            Fixed-8 (RQ-VAE) & 8.00 & 5.206 & 7.574 & 0.02605 & 0.10664 & 69.84 & 59.87 & 67.88 / 63.78 / 54.67 \\
            L2 argmin (ARQ-VAE) & 3.55 & \textbf{3.152} & \textbf{5.521} & \textbf{0.02068} & \textbf{0.09296} & \textbf{74.92} & \textbf{62.01} & \textbf{69.20} / \textbf{65.18} / \textbf{57.59} \\
            L2-permuted & 3.55 & \textbf{3.152} & \textbf{5.521} & 0.02380 & 0.09900 & 70.50 & 58.98 & 64.41 / 60.81 / 53.15 \\
            Label-aware probe & 5.13 & 3.880 & 6.249 & 0.02420 & 0.10087 & 68.45 & 51.11 & 57.52 / 54.10 / 44.57 \\
            \bottomrule
        \end{tabular}}
    \end{table}

    The cleanest controlled comparison in~\cref{tab:selection-criteria} is L2 argmin versus L2-permuted: both use the same depth histogram over all frames and 3.152~MB of storage, yet preserving the reconstruction-derived frame/depth association improves F1@50 from 53.15 to 57.59 and lowers MSE from 0.02380 to 0.02068. The gain is therefore not explained by the average number of retained codes alone; assigning deeper prefixes to the right frames matters. The label-aware probe uses TAS supervision and a task-discriminative loss, but it retains deeper prefixes (5.13 on average) and does not improve selection in this experiment. We do not claim that feature-space L2 is a universal segmentation objective; these results support it as a useful label-free criterion that yields effective frame-specific allocation.

    \paragraph{Explicit rate--distortion control.}
    L2 argmin minimizes distortion without an explicit storage penalty. We therefore also select prefixes with the rate--distortion objective
    \begin{equation}
        m_t(\lambda) = \argmin_{m \in \{1,\dots,M\}} \big[ D_t(m) + \lambda m \big],
        \label{eq:rd_selection}
    \end{equation}
    where $D_t(m)$ is the reconstruction MSE of the length-$m$ prefix. Since every stage stores one index from a $K{=}128$ codebook, $m$ is proportional to the index rate. All points reuse the same RQ-VAE and cached prefix errors, so no allocator or additional codec training is introduced. Positive $\lambda$ penalizes rate, $\lambda{=}0$ recovers L2 argmin, and negative values are included as high-rate diagnostics that favor additional stages.

    \begin{table}[h]
        \centering
        \small
        \setlength{\tabcolsep}{5pt}
        \caption{\textbf{Rate--distortion prefix selection},~\cref{eq:rd_selection}, on 50Salads with MS-TCN. Storage excludes the decoder; Complete includes it.}
        \label{tab:rd-sweep}
        \resizebox{\linewidth}{!}{\begin{tabular}{lccccccc}
            \toprule
            Criterion & Mean depth & Storage (MB) & Complete (MB) & MSE $\downarrow$ & Acc & Edit & F1@50 \\
            \midrule
            Fixed-8 (RQ-VAE) & 8.00 & 5.206 & 7.574 & 0.02605 & 69.84 & 59.87 & 54.67 \\
            $\lambda{=}10^{-3}$ (lower rate) & 2.06 & \textbf{2.466} & \textbf{4.835} & 0.02126 & 68.54 & 53.75 & 50.65 \\
            $\lambda{=}10^{-4}$ (mild rate penalty) & 3.30 & 3.037 & 5.406 & 0.02069 & 72.64 & 57.87 & 56.05 \\
            $\lambda{=}0$ (L2 argmin, ARQ-VAE) & 3.55 & 3.152 & 5.521 & \textbf{0.02068} & 74.92 & \textbf{62.01} & 57.59 \\
            $\lambda{=}{-}10^{-4}$ (mild high-rate bias) & 3.82 & 3.277 & 5.646 & 0.02069 & \textbf{75.28} & 60.57 & \textbf{58.21} \\
            $\lambda{=}{-}10^{-3}$ (strong high-rate bias) & 6.15 & 4.353 & 6.722 & 0.02191 & 74.67 & 56.75 & 54.07 \\
            \bottomrule
        \end{tabular}}
    \end{table}

    As shown in~\cref{tab:rd-sweep}, positive $\lambda$ provides explicit lower-rate control, reducing mean depth and storage with the expected performance trade-off. A mild high-rate bias at $\lambda{=}{-}10^{-4}$ increases mean depth only from 3.55 to 3.82 and complete storage from 5.521 to 5.646~MB, while marginally improving F1@50 from 57.59 to 58.21. Thus, when the storage budget permits, $\lambda<0$ can select a slightly higher-rate operating point with improved downstream performance. A stronger bias at $\lambda{=}{-}10^{-3}$ instead reduces F1@50 to 54.07, showing that retaining more stages is not monotonically beneficial. The $\lambda$ term therefore provides controllable operating points for different storage budgets.

    \paragraph{Why can a shorter prefix reconstruct better?}
    A standard RQ-VAE is optimized for full-depth reconstruction, but pointwise monotonic improvement is not guaranteed once residual corrections learned in latent space are passed through a nonlinear decoder. This is especially relevant for heterogeneous long-video features spanning many scenes, actions, and levels of visual complexity: many frames lie in visually stable action interiors, while others capture transitions, viewpoint changes, or complex motion, and for frames already represented well by a short prefix, later stages may add unnecessary detail in feature space. This behavior does not stem from unstable optimization: in our training, reconstruction loss decreases stably while codebook perplexity increases steadily, with no indication of codebook collapse. Nor is ARQ-VAE merely deleting globally harmful late stages. The redundancy analysis in Appendix~\ref{app:prefix_difficulty} finds systematically shorter prefixes for temporally redundant frames and longer prefixes for harder-to-reconstruct regions, and the L2-permuted control above shows that \emph{which} frames receive deeper prefixes matters. Consistent with this explanation, ARQ-VAE lowers MSE from 0.02605 to 0.02068 and improves F1@50 from 54.67 to 57.59 relative to Fixed-8 RQ-VAE.

    \paragraph{Selecting prefixes before fine-tuning.}
    Fine-tuning the decoder and active codebooks in ARQ-VAE$^*$ can change which prefix minimizes reconstruction error. ARQ-VAE$^*$ deliberately fine-tunes with the selected indices and prefix lengths held fixed, directly adapting the model to the compact representation that will actually be stored. This preserves the stored payload and avoids an alternating procedure in which the selection, and therefore the storage, changes after every update.

    \section{Computational Cost}
    \label{app:compute}

    All computational-cost results in this section use the \textbf{40-video I3D training feature bank of 50Salads}, with the $M{=}8$, $K{=}128$ codec and a 50-epoch downstream MS-TCN schedule. These timings describe a single training run. We report codec-related compute separately from downstream MS-TCN optimization. Our protocol decodes compressed features on the fly, so each epoch reconstructs each video batch from its stored codes. When sufficient transient memory is available, the compact bank may alternatively be decoded once before TAS optimization, trading memory residency for reduced repeated decoding; all timings below use on-the-fly reconstruction. All timings were measured on a single NVIDIA Tesla V100-SXM2 GPU with 32~GB of memory, using PyTorch 2.6.0 with CUDA 11.8 and Python 3.10.18; absolute times will vary across hardware.

    \begin{table}[h]
        \centering
        \small
        \setlength{\tabcolsep}{6pt}
        \caption{\textbf{Codec-related computational cost} on 50Salads. Codec preparation is a one-time offline cost; decoding reconstructs the complete 40-video training bank once per epoch, and the 50-epoch columns follow the MS-TCN schedule.}
        \label{tab:compute}
        \begin{tabular}{lcccc}
            \toprule
            Representation & Codec prep (s) & Decode / epoch (s) & Decode, 50 epochs (s) & Codec total (s) \\
            \midrule
            RQ-VAE & \textbf{102.91} & 3.78 & 189.00 & 291.91 \\
            ARQ-VAE & 119.38 & \textbf{3.43} & \textbf{171.50} & \textbf{290.88} \\
            ARQ-VAE$^*$ & 333.49 & 3.54 & 177.00 & 510.49 \\
            \bottomrule
        \end{tabular}
    \end{table}

    \begin{figure}[h]
        \centering
        \includegraphics[width=\linewidth]{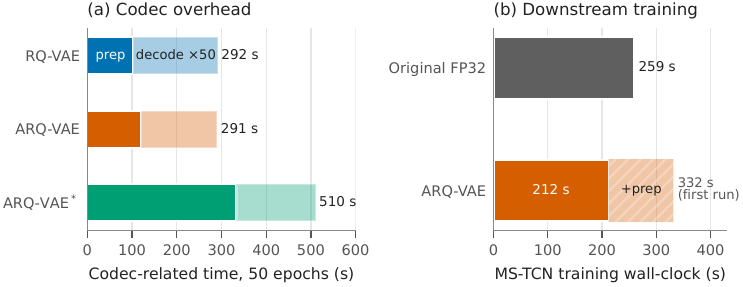}
        \caption{\textbf{Computational cost} on 50Salads. \textbf{(a)}~Codec-related time over the 50-epoch MS-TCN schedule: one-time preparation (solid) plus on-the-fly decoding (light). \textbf{(b)}~End-to-end MS-TCN training wall-clock, including feature access and on-the-fly reconstruction; the hatched segment adds ARQ-VAE's one-time preparation, paid only on the first run.}
        \label{fig:compute}
    \end{figure}

    For RQ-VAE, codec preparation in~\cref{tab:compute} includes codec training and code extraction. ARQ-VAE adds prefix reconstruction, error evaluation, and selection, an incremental cost of only 16.47~s. ARQ-VAE$^*$ additionally fine-tunes the decoder and active codebooks, increasing preparation time from 119.38~s to 333.49~s (about $2.8\times$) while leaving decoding time nearly unchanged. ARQ-VAE reconstructs the complete bank in 3.43~s per epoch, slightly faster than fixed-depth RQ-VAE because it decodes fewer residual stages on average. The codec totals in~\cref{tab:compute} include preparation and repeated decoding, but exclude MS-TCN optimization; decoding is already included in the end-to-end training times and must not be added again. As shown in~\cref{fig:compute}, under the same 50-epoch schedule, end-to-end MS-TCN fitting took 212.30~s with ARQ-VAE versus 259.43~s with the original FP32 bank, so decoding did not increase the observed downstream training wall-clock in this run. Including the one-time 119.38~s preparation, the first ARQ-VAE run costs 331.68~s; subsequent runs on the same compact bank reuse this preparation.

    \section{Generalization to Alternative Pre-extracted Features}\label{app:dino}
    In~\cref{tab:dinov3-results}, we evaluate whether ARQ-VAE generalizes beyond the I3D feature space by repeating the 50Salads experiments with DINOv3~\cite{simeoni2025dinov3} frame-wise features. The goal of this analysis is not to optimize a new backbone-specific pipeline, but to test whether adaptive residual quantization remains effective when the input feature distribution changes. The results show the same overall pattern observed with I3D: RQ-VAE provides a compact fixed-rate representation, ARQ-VAE reduces storage by truncating unnecessary residual stages, and ARQ-VAE$^*$ further improves reconstruction quality after fine-tuning on the retained prefixes. This suggests that the benefit of adaptive residual depth is not specific to I3D, but reflects a more general property of long-video feature banks: many frames can be represented with shallow residual prefixes, while harder frames benefit from additional codes.

    \begin{table}[t]
    \centering
    \small
    \setlength{\tabcolsep}{5pt}
    \renewcommand{\arraystretch}{1.15}
    \caption{Results on DINOv3 frame features under the same condensation pipeline. Lower is better for reconstruction metrics and storage; higher is better for TAS metrics. The results show that the proposed method is not specific to I3D features and extends effectively to a different pretrained feature space.}
    \label{tab:dinov3-results}
    \begin{tabular}{lcccccccc}
        \toprule
        \multirow{2}{*}{Method}
        & \multicolumn{3}{c}{Reconstruction}
        & \multicolumn{5}{c}{TAS metrics} \\
        \cmidrule(lr){2-4} \cmidrule(lr){5-9}
        & MSE $\downarrow$ & MAE $\downarrow$ & Storage $\downarrow$
        & Acc $\uparrow$ & Edit $\uparrow$ & F1@10 $\uparrow$ & F1@25 $\uparrow$ & F1@50 $\uparrow$ \\
        \midrule
        Original
        & -- & -- & 4.5 GB
        & \textbf{70.4} & \textbf{67.9} & \textbf{69.6} & \textbf{66.8} & \textbf{54.3} \\
        RQ-VAE
        & 0.034 & 0.143 & 4.1 MB
        & 60.3 & 53.9 & 59.1 & 55.8 & 41.9 \\
        ARQ-VAE
        & 0.017 & 0.099 & \textbf{1.8} MB
        & 65.1 & 56.6 & 57.9 & 53.2 & 41.1 \\
        ARQ-VAE$^{*}$
        & \textbf{0.014} & \textbf{0.092} & \textbf{1.8} MB
        & 67.2 & 61.6 & 63.8 & 57.9 & 45.2 \\
        \bottomrule
    \end{tabular}
\end{table}

    \section{Results with a Recent TAS Backbone (FACT)}
    \label{app:fact}

    ARQ-VAE modifies only the stored training feature bank; the downstream TAS architecture, its inference-time input, and its training objective are unchanged, so it is compatible in principle with any feature-based TAS model. To verify this with a more recent architecture, we integrate FACT~\cite{lu2024fact} into our codebase and run the complete 5-split 50Salads cross-validation for four training banks. Every row uses the same splits, seeds, codec checkpoints, and code pools as our MS-TCN experiments, so the segmentation backbone is the only change; each compressed row trains FACT on the reconstructed feature bank and evaluates it on the original test features.

    \begin{table}[h]
        \centering
        \small
        \setlength{\tabcolsep}{8pt}
        \caption{\textbf{FACT~\cite{lu2024fact} on 50Salads}, averaged over the five standard splits. Original trains on the full I3D feature bank; the other rows train on reconstructed features.}
        \label{tab:fact}
        \begin{tabular}{lccccc}
            \toprule
            Method & Acc & Edit & F1@10 & F1@25 & F1@50 \\
            \midrule
            Original & 85.10 & 79.29 & 84.77 & 83.08 & 77.61 \\
            \midrule
            RQ-VAE & 77.65 & 70.99 & 76.30 & 74.16 & 66.28 \\
            ARQ-VAE & 79.13 & 73.02 & 79.06 & 76.30 & 66.52 \\
            ARQ-VAE$^*$ & \textbf{80.23} & \textbf{73.38} & \textbf{79.26} & \textbf{77.19} & \textbf{67.28} \\
            \bottomrule
        \end{tabular}
    \end{table}

    \begin{figure}[h]
        \centering
        \includegraphics[width=0.62\linewidth]{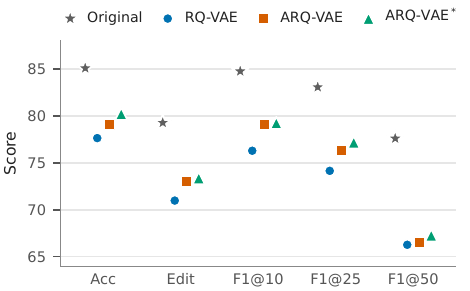}
        \caption{\textbf{FACT on 50Salads} (5-split average). The ordering RQ-VAE $<$ ARQ-VAE $<$ ARQ-VAE$^*$ observed with MS-TCN and ASFormer is unchanged.}
        \label{fig:fact}
    \end{figure}

    As shown in~\cref{tab:fact} and~\cref{fig:fact}, ARQ-VAE$^*$ is the best of the three stored representations on all five metrics, and it outperforms RQ-VAE in all five splits on Acc, F1@10, F1@25, and F1@50 (+2.58 Acc and +2.96 F1@10 on average). The ordering RQ-VAE $<$ ARQ-VAE $<$ ARQ-VAE$^*$ is unchanged from MS-TCN and ASFormer, and swapping MS-TCN for FACT required no change to the method. Moreover, FACT trained on the 2.4~MB ARQ-VAE$^*$ memory exceeds MS-TCN trained on the full 4.5~GB feature bank (\cref{tab:main-tas}) on all five metrics, showing that a stronger architecture and the compact feature bank are complementary.
    
\fi

\newpage

\newif\ifshowchecklist

\showchecklistfalse

\ifshowchecklist
\section*{NeurIPS Paper Checklist}

\begin{enumerate}

\item {\bf Claims}
    \item[] Question: Do the main claims made in the abstract and introduction accurately reflect the paper's contributions and scope?
    \item[] Answer: \answerYes{} 
    \item[] Justification: All claims in the abstract and introduction are aligned with paper's contribution both experimentally and methodologically. Contributions are outlined in both abstract and introduction.

\item {\bf Limitations}
    \item[] Question: Does the paper discuss the limitations of the work performed by the authors?
    \item[] Answer: \answerYes{} 
    \item[] Justification: Limitations are mentioned throughout the experimental discussion and in the conclusion section.

\item {\bf Theory assumptions and proofs}
    \item[] Question: For each theoretical result, does the paper provide the full set of assumptions and a complete (and correct) proof?
    \item[] Answer: \answerNA{} 
    \item[] Justification: We do not include any theoretical results in this paper.

    \item {\bf Experimental result reproducibility}
    \item[] Question: Does the paper fully disclose all the information needed to reproduce the main experimental results of the paper to the extent that it affects the main claims and/or conclusions of the paper (regardless of whether the code and data are provided or not)?
    \item[] Answer: \answerYes{} 
    \item[] Justification: Fully usable and reproducable code is included in the supplementary. Code is easy to run and contains a clear README file explaining the steps for reproduction.

\item {\bf Open access to data and code}
    \item[] Question: Does the paper provide open access to the data and code, with sufficient instructions to faithfully reproduce the main experimental results, as described in supplemental material?
    \item[] Answer: \answerYes{} 
    \item[] Justification: All code is included in the supplementary, and all datasets are standard benchmarks that are both publicly available and used often in this field of research.

\item {\bf Experimental setting/details}
    \item[] Question: Does the paper specify all the training and test details (e.g., data splits, hyperparameters, how they were chosen, type of optimizer) necessary to understand the results?
    \item[] Answer: \answerYes{} 
    \item[] Justification: All implementation details are given in the experimental setup part of experimental section 4. 

\item {\bf Experiment statistical significance}
    \item[] Question: Does the paper report error bars suitably and correctly defined or other appropriate information about the statistical significance of the experiments?
    \item[] Answer: \answerNo{} 
    \item[] Justification: We do not explicitly provide statistical significance tests, although is not common in the field of Temporal Action Segmentation. We use full cross validation on all datasets for all main experiments (detailed in experimental section), and multiple task sequences for incremental TAS.

\item {\bf Experiments compute resources}
    \item[] Question: For each experiment, does the paper provide sufficient information on the computer resources (type of compute workers, memory, time of execution) needed to reproduce the experiments?
    \item[] Answer: \answerYes{} 
    \item[] Justification: All results are rather quick to train, as we work with small neural networks that are detailed in the experiment setup. Also, we give brief GPU information in the experimental details. 
    
\item {\bf Code of ethics}
    \item[] Question: Does the research conducted in the paper conform, in every respect, with the NeurIPS Code of Ethics \url{https://neurips.cc/public/EthicsGuidelines}?
    \item[] Answer: \answerYes{} 
    \item[] Justification: All research is conducted according to NeurIPS Code of Ethics.

\item {\bf Broader impacts}
    \item[] Question: Does the paper discuss both potential positive societal impacts and negative societal impacts of the work performed?
    \item[] Answer: \answerYes{} 
    \item[] Justification: See Appendix~\ref{app:broader} for discussion on positive and negative broader societal impacts.
    
\item {\bf Safeguards}
    \item[] Question: Does the paper describe safeguards that have been put in place for responsible release of data or models that have a high risk for misuse (e.g., pre-trained language models, image generators, or scraped datasets)?
    \item[] Answer: \answerNA{} 
    \item[] Justification: 

\item {\bf Licenses for existing assets}
    \item[] Question: Are the creators or original owners of assets (e.g., code, data, models), used in the paper, properly credited and are the license and terms of use explicitly mentioned and properly respected?
    \item[] Answer: \answerYes{} 
    \item[] Justification: All code, models, and data used from external owners are cited in our paper.

\item {\bf New assets}
    \item[] Question: Are new assets introduced in the paper well documented and is the documentation provided alongside the assets?
    \item[] Answer: \answerYes{} 
    \item[] Justification: All information regarding anonymized code is given in the README document.

\item {\bf Crowdsourcing and research with human subjects}
    \item[] Question: For crowdsourcing experiments and research with human subjects, does the paper include the full text of instructions given to participants and screenshots, if applicable, as well as details about compensation (if any)? 
    \item[] Answer: \answerNA{} 
    \item[] Justification: 

\item {\bf Institutional review board (IRB) approvals or equivalent for research with human subjects}
    \item[] Question: Does the paper describe potential risks incurred by study participants, whether such risks were disclosed to the subjects, and whether Institutional Review Board (IRB) approvals (or an equivalent approval/review based on the requirements of your country or institution) were obtained?
    \item[] Answer: \answerNA{} 
    \item[] Justification: 

\item {\bf Declaration of LLM usage}
    \item[] Question: Does the paper describe the usage of LLMs if it is an important, original, or non-standard component of the core methods in this research? Note that if the LLM is used only for writing, editing, or formatting purposes and does \emph{not} impact the core methodology, scientific rigor, or originality of the research, declaration is not required.
    \item[] Answer: \answerNA{} 
    \item[] Justification: 

\end{enumerate}
\fi

\end{document}